\documentclass{article}

\PassOptionsToPackage{numbers, compress}{natbib}

\usepackage[main, final]{neurips_2026}
\usepackage[T1]{fontenc}    
\usepackage{url}            
\usepackage{booktabs}       
\usepackage{amsfonts}       
\usepackage{nicefrac}       
\usepackage{microtype}      
\usepackage{graphicx} 
\usepackage{multirow}
\usepackage{array}
\usepackage{amsmath}
\usepackage{xcolor}
\newcolumntype{L}{>{$}l<{$}}
\newcolumntype{R}{>{$}r<{$}}
\newcolumntype{C}{>{$}c<{$}}
\usepackage[detect-all,separate-uncertainty = true]{siunitx}
\usepackage{fontawesome5}
\usepackage{colortbl}
\usepackage[dvipsnames]{xcolor}
\definecolor{citeblue}{rgb}{0.21,0.49,0.74}
\usepackage[pagebackref,breaklinks,colorlinks,citecolor=citeblue,linkcolor=Blue]{hyperref}       %
\usepackage{xfakebold}
\usepackage{arydshln}
\usepackage{subcaption}
\usepackage{float}
\usepackage{circledsteps}
\usepackage[inline]{enumitem}
\usepackage{wrapfig}
\usepackage[most]{tcolorbox}

\definecolor{realgreen}{RGB}{94, 107, 71}
\definecolor{randomyellow}{RGB}{253, 185, 51}
\definecolor{removegrey}{RGB}{212, 212, 212}
\definecolor{oursred}{RGB}{206, 80, 47}

\newcommand{\fbseries}{\unskip\setBold\aftergroup\unsetBold\aftergroup\ignorespaces}
\makeatletter
\newcommand{\setBoldness}[1]{\def\fake@bold{#1}}
\makeatother

\usepackage{xspace}
\newcommand{\latent}{\texttt{<latent>}\xspace}
\newcommand{\latentstart}{\texttt{<latent\_start>}\xspace}
\newcommand{\latentend}{\texttt{<latent\_end>}\xspace}

\title{Leveraging Latent Visual Reasoning in Silence}

\author{
  \bfseries Dongyao Zhu$^{1\dagger}$ \quad Zhen Wang$^3$ \quad Xi Xiao$^4$ \quad Han Jiang$^5$ \quad Saeed Vahidian$^6$ \\
  \bfseries Wei-Lun Chao$^7$ \quad Tanya Berger-Wolf$^2$ \quad Yu Su$^2$ \quad Raju Vatsavai$^1$ \quad Jianyang Gu$^{2\dagger}$ \\[0.5em]
  \normalfont\small $^1$North Carolina State University \quad $^2$The Ohio State University \quad $^3$UC, San Diego \\ $^4$\small UAB \quad \small $^5$JHU \quad $^6$Duke University \quad $^7$Boston University \\
  {\small $\dagger$ \texttt{dzhu6@ncsu.edu} \quad \texttt{gu.1220@osu.edu}}
}

\begin{document}

\maketitle

\begin{figure}[h]
    \centering
    \vspace{-16pt}
    \includegraphics[width=\linewidth]{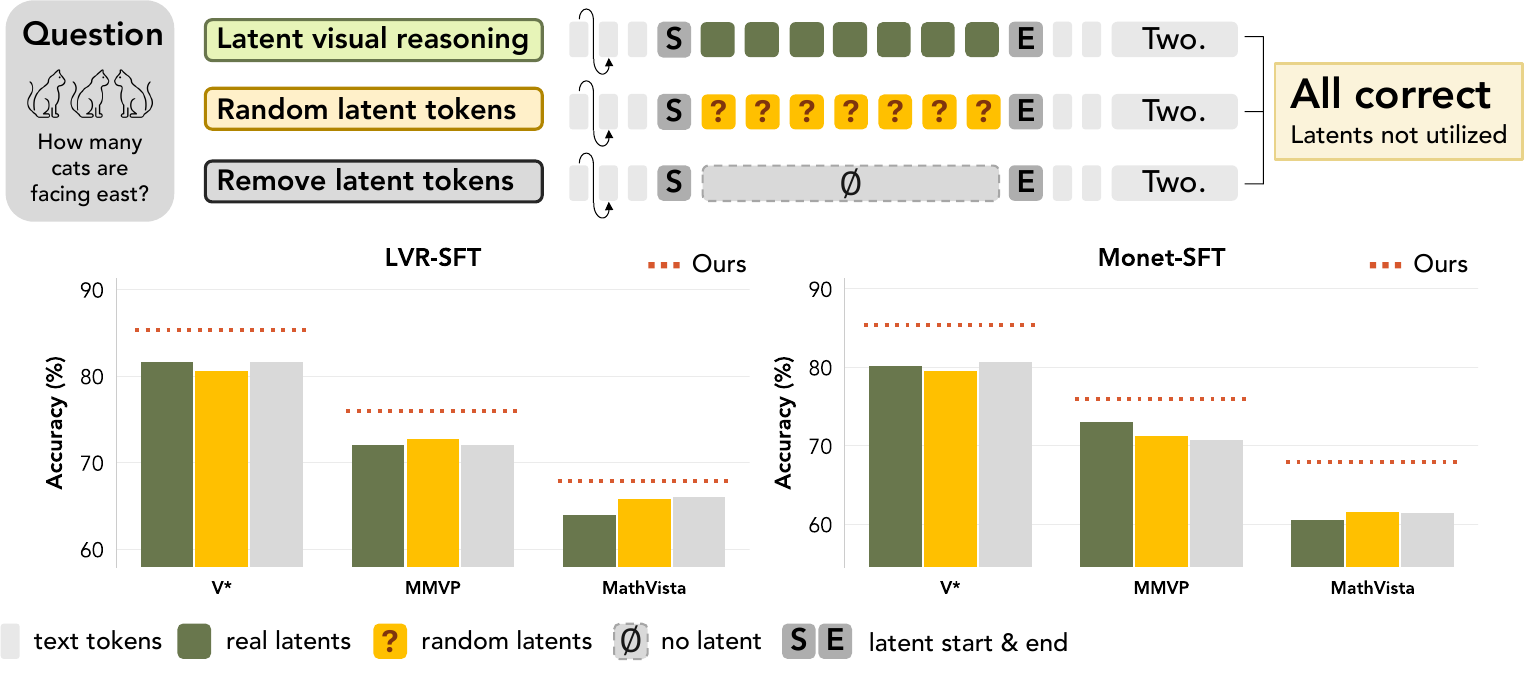}
    \vspace{-12pt}
    \caption{When standard latent tokens (\textcolor{realgreen}{green}) are replaced with random values (\textcolor{randomyellow}{yellow}) or completely removed (\textcolor{removegrey}{grey}), latent visual reasoning models retain comparable or achieve even higher performance across benchmarks. We demonstrate that latent reasoning may have limited value as a mandatory inference-time format, but can still be leveraged \textbf{in silence} to improve learning (\textcolor{oursred}{red}).}
    \label{fig:intro_0}
\end{figure}

\begin{abstract}
    Latent visual reasoning involves visual evidence more directly in multimodal reasoning by inserting continuous latent tokens before textual generation. However, the necessity of these latent tokens at inference remains ambiguous. We show that replacing latent tokens with random noise or removing them completely causes little performance degradation across spatial reasoning benchmarks. Reinforcement learning further diminishes the latent generation behavior after post-training. These observations raise a central question: \textit{Is latent visual reasoning still meaningful}? We argue that its value should be measured by how effectively latent tokens guide learning, rather than whether they persist as an inference-time format. Our analysis shows that latent reasoning is unevenly favorable across question types, yet hard task-level routing for applying latent generation is brittle. Motivated by these findings, we propose an attention-based reward that encourages generated latent tokens to interact with later text tokens during RL. This reward promotes latent utilization when the latent mode is activated while preserving the flexibility to use pure-text reasoning. Experiments show that our method improves performance across perception and visual reasoning benchmarks, even when latent tokens are rarely generated after post-training. Our results highlight that, without explicit expression at inference, latent visual reasoning can shape better visual grounding and more accurate textual reasoning \textbf{in silence}. 
Code \& model: \href{https://github.com/ddydyd32/silent-lvr/tree/master}{\faGithub}.

\end{abstract}

\section{Introduction}
Latent visual reasoning offers an appealing alternative to purely text-based reasoning in multimodal models~\cite{li2025latent, wang2025monet, tong2025sketch}. This paradigm allows models to generate continuous latent embeddings before discrete textual generation. These embeddings are usually optimized to capture visual information from prompt-related regions or auxiliary visual inputs such as depth maps~\cite{li2025latent, wang2025monet}. Many visual reasoning tasks depend on spatial evidence that can be difficult to express fully in language~\cite{lu2023mathvista, thrush2022winoground, liu2024mmbench, fu2024blink, xu2025visulogic,xiao2026staying}. If properly leveraged, latent tokens could summarize visual information more directly and improve the grounding of multimodal reasoning. This work seeks an opposite direction and asks whether latent visual reasoning can be useful even when it is not preserved as an explicit inference-time behavior. We refer to this possibility as leveraging latent visual reasoning ``\textbf{in silence}.''

This view is motivated by a basic ambiguity in current latent visual reasoning models: latent generation alone does not guarantee latent usage~\cite{wang2025monet}. Building on this observation, we revisit latent dependence across representative models such as Monet~\cite{wang2025monet} and LVR~\cite{li2025latent} under perturbations to the latent tokens. Applied perturbations include replacing the generated latent tokens with random noise before subsequent text generation or removing them completely. Across popular visual reasoning benchmarks~\cite{cheng2025v, liu2024mmbench, lu2023mathvista}, \autoref{fig:intro_0} shows that these perturbations cause little performance degradation. The stable performance suggests that latent visual reasoning models may learn the format of latent generation, but rely only weakly on the generated tokens for the final answer.

This ambiguity becomes more pronounced during post-training through reinforcement learning (RL). We apply standard correctness-rewarded GRPO~\cite{guo2025deepseek} to latent visual reasoning models and find that RL tends to diminish latent reasoning. As training proceeds, models increasingly avoid generating latent tokens and eventually converge toward pure-text reasoning. 
This behavior is counterintuitive, yet aligned with our earlier latent-dependence analysis. If latent tokens do not consistently improve the final answer, an outcome-oriented reward will provide little incentive to preserve them. To test whether this collapse is due to the lack of an explicit latent-related objective, we design a diagnostic latent necessity reward that is positive only when the response with latent is more accurate than its no-latent counterpart. During RL, this reward quickly drives the model to skip latent generation for most questions while maintaining response quality (\S\ref{sec:latent-behavior}). This further suggests that latent visual tokens may have limited value as a mandatory inference-time format.

These findings raise the central question of our work: \textit{if latent tokens can often be removed without hurting performance, and RL tends to eliminate them, what is the value of latent visual reasoning}?
We argue that the focus should shift from latent presence to latent utilization, meaning latent tokens should influence subsequent textual reasoning.
Even when the final model does not always emit latent tokens, latent visual reasoning may still guide the learning process \textbf{in silence}.
To further investigate the underlying mechanism, we compare the performance of reasoning with latents and pure text across different question types. We find that latent reasoning and pure-text responses are unevenly favorable for different questions and model families (\autoref{fig:rollout_groups_text}).
While this makes identifying task type and applying corresponding objectives challenging, it motivates a more practical approach: improve the utilization of latent tokens without forcing latent generation on every example.

To this end, we propose an attention-based reward for latent visual reasoning during RL. 
Our intuition is that a useful latent token should effectively influence the tokens generated after it. We therefore use the attention from subsequent text tokens to latent tokens as an efficient proxy for latent influence over next-token prediction (\S\ref{sec:method}). When combined with RL, this reward strengthens the interaction between latent tokens and textual reasoning. Empirically, models trained with this reward attend substantially more often to generated latent tokens and achieve improved final performance, even if latent generation still decreases later in training (\S\ref{sec:result}). This outcome echoes our hypothesis that latent reasoning can improve learning in silence. We further study model behavior through attribution analyses (\S\ref{sec:app-causal}) and qualitative examples (\S\ref{sec:qualitative}) to interpret why our model performs better without generating latents.
Results suggest that latent reasoning is useful less as a persistent response format and more as a training-time scaffold that shapes the learning process.

Through this study, we transition our focus from whether latent reasoning is necessary to whether latents are appropriately leveraged. 
While the presence of latents is not a reliable indicator of better performance, the attention reward can be used as an effective post-training signal. Our results suggest that the future of latent visual reasoning depends on training objectives that make latent computation influential during learning, even when its benefits are ultimately expressed in silence.

\section{Related Work}

\textbf{Explicit reasoning in multimodal models.}
A prominent line of work improves multimodal reasoning by introducing explicit, interpretable intermediate steps.
Originally introduced in the text domain, chain-of-thought prompting~\cite{wei2022chain} demonstrates that eliciting step-by-step natural language reasoning substantially improves complex problem solving.
In the visual domain, explicit reasoning takes several forms: textual verbalization of visual content~\cite{changpinyo2022all, hu2022promptcap}, active tool use such as grounding, zooming, cropping, and contouring~\cite{zhou2025reinforced, chen2023large, qi2024cogcom, zheng2025deepeyes, bai2025univg}, and generation of new visual information via code execution or external vision models such as depth estimators and segmentation networks~\cite{hu2024visual, li2025imagine, wu2025vtool}.
A separate but related trend develops unified architectures that natively interleave text, image, and audio within a single model, typically featuring unimodal tokenizers~\cite{sun2024multimodal, team2024chameleon,li2025imagine}. In contrast to these approaches that externalize reasoning into human-interpretable modalities, our work studies reasoning in latent space, where intermediate computation is not directly observable.

\textbf{Latent visual reasoning and reinforcement learning.}
Recent work also explores reasoning in continuous latent space. Some approaches feed predicted embeddings directly as inputs for subsequent steps~\cite{hao2024training}. Others generate free-form latent representations before decoding~\citep{shen2025codi, tong2025sketch}. A complementary direction constrains latent computation to a finite set of abstract tokens~\cite{ramji2026thinking}.
In the multimodal setting, the output hidden states are usually trained to carry related visual information~\cite{li2025latent, wang2025monet} before interleaving with text. These representations are often referred to as visual latents.
Subsequent work improves the helpfulness of visual latents via task-agnostic training~\citep{li2025latent} or by coupling visual latents with external vision foundation models ~\citep{qin2025chain}. 
Despite these efforts, these methods implicitly assume that latent representations are both generated and meaningfully utilized during reasoning, especially in model post-training, which is typically based on RL~\cite{guo2025deepseek}. 
Standard correctness/format rewards supervise final answers but not intermediate steps.
While Monet~\citep{wang2025monet} addresses this issue with their VLPO variant, we show that optimizing for latent generation does not guarantee its actual utilization during reasoning. We note concurrent work~\cite{li2026imagination} that addresses underutilized visual latents by removing them entirely and reverting to explicit intermediate representations such as captioning. 
In contrast, we show that visual latents are not necessary at inference but can still shape learning via utilization signals \textbf{in silence}.

\section{Latent Generation is Unnecessary as an Inference-Time Behavior}

Latent visual reasoning augments multimodal language models with continuous latent embeddings for intermediate visual computation~\cite{li2025latent, wang2025monet, tong2025sketch}. Instead of expressing the entire reasoning process through discrete language tokens, a latent visual reasoning model can switch into a latent mode by generating a special signal token, such as \latentstart, producing a span of continuous \latent tokens, and then return to textual generation after \latentend. These latent tokens are usually designed to encode task-relevant visual information, either from prompt-related regions in the original image or from auxiliary visual inputs such as depth maps~\cite{li2025latent,wang2025monet}. The resulting response contains both ordinary text tokens and latent tokens. 
Previous methods produce latent tokens that improve performance on visual reasoning tasks over text-only baselines.
This section investigates the behavior of these models during inference and after RL post-training.

\subsection{Latent Dependence During Inference}\label{sec:latent_dependence}
We first examine whether the latent visual reasoning models actually rely on their generated tokens during inference. Specifically, we apply perturbations to the generated \latent tokens and evaluate the answer correctness. This question is central to understanding the role of these latents, which are expected to provide grounded visual information for subsequent textual reasoning.
To isolate the effect from post-training dynamics, we conduct this analysis on latent reasoning models trained with supervised fine-tuning (SFT) before RL. All experiments use a temperature $\approx 0.0$, following common evaluation settings~\citep{duan2024vlmevalkit}. We apply the following two perturbations:

\noindent \textbf{Random replacement.}
Models can generate $k$ \latent tokens before \latentend. However, before the subsequent generation step, we replace latent tokens with random values drawn from a unit Gaussian distribution. Positional embeddings are kept unchanged. We set $k$ to 8 for LVR and 10 for Monet, following the settings used in their best-reported performance. 

\noindent \textbf{Removing latent tokens.}
In this setting, models can still predict \latentstart, but we force the token \latentend as the next token, skipping the continuous latent input. Compared with random replacement, this setting tests whether the presence of \latent tokens is needed at all.

As shown in \autoref{fig:intro_0}, we observe that random replacement performs on par with model-generated latent tokens. Even though removing latent tokens is out of the training distribution, it is better than using them in many cases. These results suggest that \latent tokens can often be removed without hurting textual reasoning. In other words, the generated tokens are not consistently necessary for producing the final answer. However, these observations do not imply that latent visual reasoning should be discarded.
First, in certain cases, such as MMVP~\cite{zhang2024mmvp}, responses still benefit from using visual latent tokens. Second, prior work reports that models trained under the latent visual reasoning paradigm outperform those trained with pure textual reasoning on the same data~\cite{wang2025monet,li2025latent}. 
Although not universally required at inference, latent visual reasoning may still provide value in specific contexts or during training. Motivated by this distinction, we further investigate the latent visual reasoning behavior after different training stages.

\subsection{Latent Generation Behavior}\label{sec:latent-behavior}
We examine whether models prefer latent generation to pure-text reasoning across different training stages, particularly after RL post-training. This analysis complements our previous perturbation study. If latent tokens are not consistently necessary for final-answer accuracy, an outcome-oriented RL objective may gradually reduce the use of latent mode. We therefore measure how often latent generation is triggered in SFT models and RL-post-trained models.

\begin{table*}[t]
    \centering
    \vspace{-16pt}
    \caption{The ratio (\%) of latent generation triggered during inference, and the accuracy (\%) of responses with latent generated or with pure text on the ViRL39K and Thyme-RL datasets. RL significantly reduces the ratio where latent mode is triggered.}
    \label{tab:rollout_stats}
    \small
    \begin{tabular}{@{}l C C C C C C C@{}}
        \toprule
        \multirow{2}{*}{\textbf{Model}} 
        & \multirow{2}{*}{\textbf{Temp.}} 
        & \multicolumn{3}{c}{\textbf{ViRL39K}} 
        & \multicolumn{3}{c}{\textbf{Thyme-RL}} \\
        \cmidrule(lr){3-5} \cmidrule(lr){6-8}
        & 
        & \text{Latent Ratio} & \text{Latent Acc} & \text{Text Acc} 
        & \text{Latent Ratio} & \text{Latent Acc} & \text{Text Acc} \\
        \midrule

        \multirow{2}{*}{LVR-SFT$^*$}  
        & 0.5 & 34.3 & 40.2 & 41.2 & 33.4 & 36.6 & 40.1 \\
        & 1.0 & 37.5 & 31.4 & 33.4 & 37.5 & 30.5 & 35.2 \\
        \midrule

        \multirow{2}{*}{Monet-SFT} 
        & 0.5 & 36.1 & 42.3 & 41.1 & 69.2 & 36.4 & 40.0 \\
        & 1.0 & 18.9 & 42.6 & 34.8 & 40.7 & 34.4 & 34.3 \\
        \midrule

        \multirow{2}{*}{Monet-RL}                   
        & 0.5 & 10.0 & 60.8 & 41.1 & 55.0 & 30.9 & 36.7 \\
        & 1.0 & 5.4  & 53.5 & 35.7 & 32.4 & 28.7 & 32.2 \\
        \bottomrule
    \end{tabular}
    \vspace{-12pt}
\end{table*}

For Monet~\cite{wang2025monet}, we directly use the released SFT and RL models. For LVR~\cite{li2025latent}, we use a retrained variant as the released LVR-SFT model is strongly constrained to follow the latent-reasoning format and always enters latent mode. To reduce this format bias, we add an additional SFT stage for $200$ steps using the Visual-COT~\cite{shao2024visual} dataset with latent targets randomly dropped. This exposes the LVR model to both latent and pure-text formats and allows it to choose between them during inference while maintaining performance. We denote the resulting model as LVR-SFT$^*$ hereafter. The detailed training information and model performance of LVR-SFT$^*$ are listed in \S\ref{sec:app-lvr-sft}.

We randomly sample $1,000$ questions from Thyme-RL~\cite{zhang2025thyme} and ViRL39K~\cite{wang2025vl}. For each question, we collect $8$ responses from each model at temperatures $0.5$ and $1.0$, and summarize the ratio of triggering latent generation and the accuracy.
As shown in \autoref{tab:rollout_stats}, 10 out of 12 settings yield a latent-mode trigger rate below $50$\%, and the RL model shows a clear trend toward pure-text reasoning. We hypothesize that RL optimization encourages models to move away from the latent-reasoning format when latent tokens do not consistently improve the final answer.
As a result, Monet-RL retains only a small amount of latent generation that leads to improved accuracy.

\textbf{Training dynamics.}
We further reproduce RL post-training to study how this latent diminishing behavior emerges. Specifically, we train Monet-SFT on the Thyme-RL dataset using VLPO~\cite{wang2025monet}. Due to computational constraints, this empirical study uses lighter configurations than those officially reported, as detailed in \S\ref{sec:rl-post-training}.

\begin{wrapfigure}{r}{0.6\textwidth}
\vspace{-18pt}
    \centering
    \includegraphics[width=\linewidth]{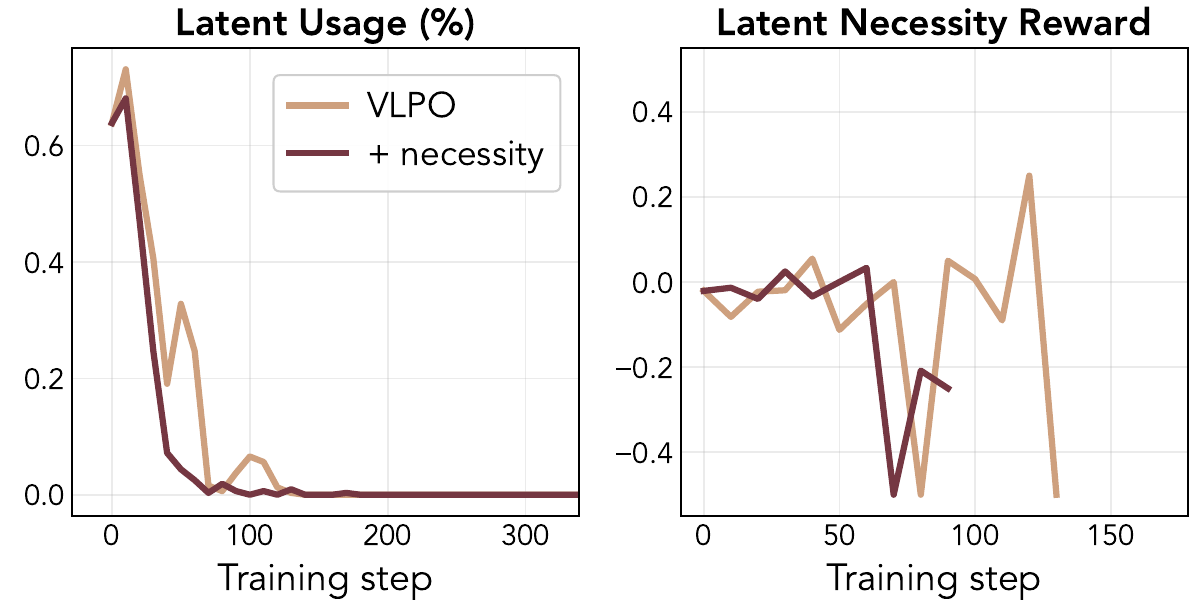}
    \caption{\label{fig:necessity} After adding the necessity reward, the model discards latent generation even faster than the VLPO baseline, without an increase in the reward value.}
\vspace{-10pt}
\end{wrapfigure}

As shown in the first column of \autoref{fig:necessity}, the ratio of responses that trigger latent mode decreases as RL training progresses. The reasonably high initial trigger ratio indicates that the model can enter latent mode at the start of training. The subsequent decrease therefore cannot be explained solely by insufficient exploration of the latent mode. A more plausible possibility is that latent-mode responses are less favorable under the RL objective.

To make this interpretation more explicit and to avoid the potential reason of lacking latent-specific rewards, we further evaluate latent necessity during RL. We define latent tokens as \textit{necessary} when keeping them yields a more accurate final response than skipping them from the same prefix before \latentstart. Based on this definition, we introduce a diagnostic latent necessity reward that is positive only when the response with generated latents is more accurate. More details on the latent necessity calculation and resulting model performance can be found in \S\ref{sec:app-necessity}. As shown in \autoref{fig:necessity} (smoothed every $10$ steps), this reward drives the model to skip latent generation even faster than vanilla RL while maintaining response quality. This result supports the conclusion that latent tokens are not consistently useful as a mandatory format at inference time. It also motivates the next step of our study: rather than preserving latent generation itself, post-training should focus on whether the latent tokens are effectively utilized.

\begin{figure}[t]
    \centering
    \vspace{-6pt}
    \includegraphics[width=0.98\linewidth]{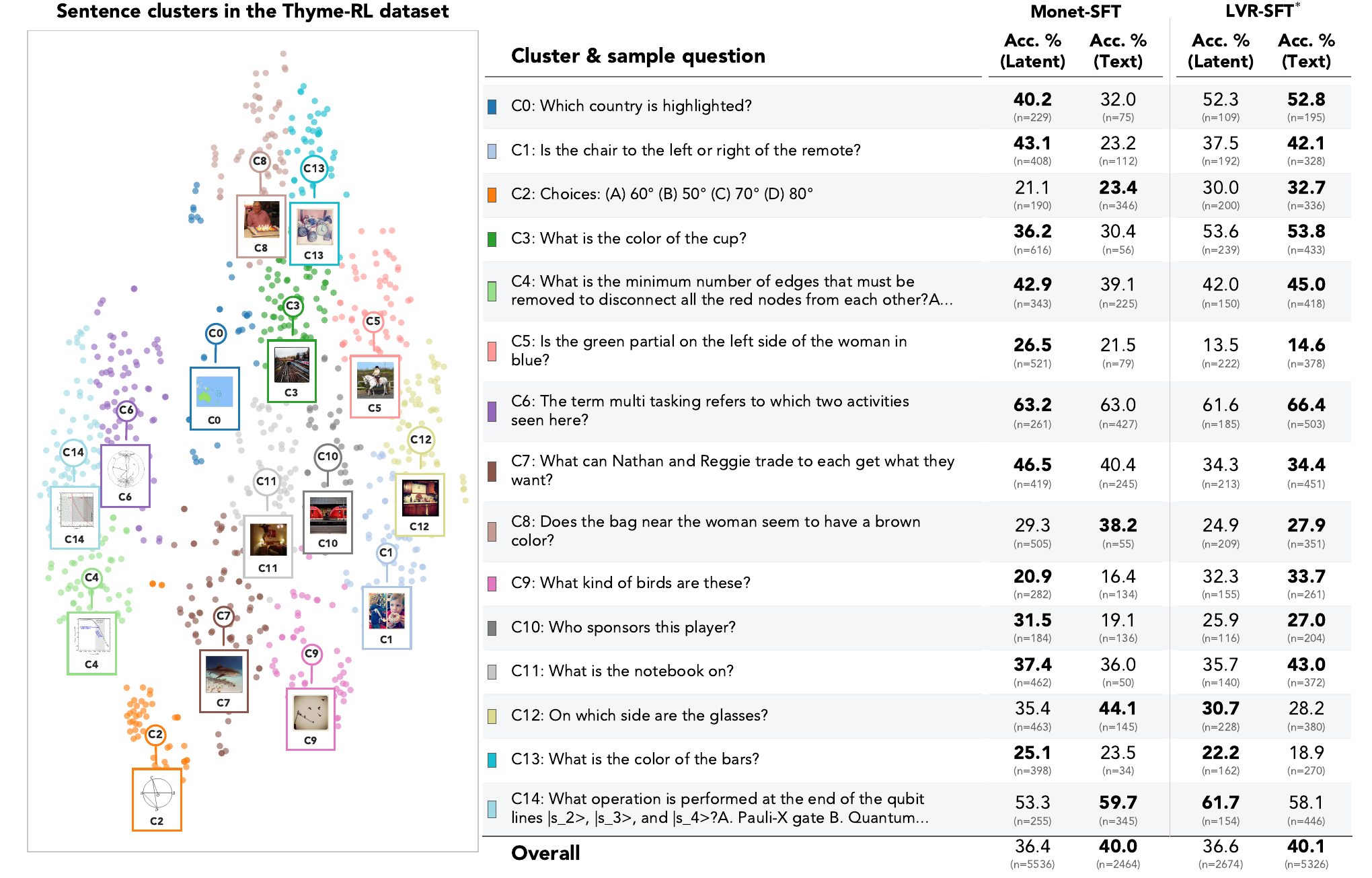}
    \caption{Accuracy comparison between using latent reasoning and text only on various question types. \textbf{Bold}: the better result between these two modes.}
    \label{fig:rollout_groups_text}
    \vspace{-12pt}
\end{figure}

\subsection{Where Does Latent Reasoning Help?}
\label{sec:rollout}

The previous section shows that RL tends to reduce latent generation, suggesting that latent tokens are not consistently useful as a mandatory inference-time format. However, this does not imply that latent reasoning is never useful. We therefore study how responses with and without latent tokens behave across different types of questions.
We construct approximate groups by clustering question embeddings using Sentence-BERT~\cite{reimers-2019-sentence-bert}. The left panel of \autoref{fig:rollout_groups_text} visualizes the resulting clusters for Thyme-RL, together with representative questions from each cluster in the middle panel. These clusters cover a mixture of question patterns, including spatial relations, object attributes, chart-like reasoning, mathematical reasoning, and general visual recognition.

The cluster-level results in \autoref{fig:rollout_groups_text} reveal a heterogeneous pattern. For Monet-SFT, latent responses outperform pure-text responses in several clusters involving local visual evidence or spatial/object attributes, whereas pure-text responses are stronger in other clusters. For LVR-SFT$^*$, the trend is different: pure-text responses are more competitive in most clusters. Therefore, a single coarse question type cannot explain latent preference, and it is not consistent across model families.

This analysis refines our interpretation of latent visual reasoning. Latent tokens can be useful, but their usefulness is conditional on various factors, including the model, the dataset, and the specific structure of the question. Some examples benefit from latent computation, while others are better handled by direct textual reasoning. This makes a hard routing of question type difficult, as deciding in advance which task category should use latent reasoning would be brittle. However, they motivate a softer post-training objective. Rather than forcing latent generation on every example or manually assigning question types to reasoning modes, we should encourage generated latent tokens to interact with subsequent textual reasoning whenever the latent mode is activated.

\section{Leveraging Latent Visual Reasoning in Silence}
\label{sec:method}
Compared with the presence of latent generation, we identify that the key issue is whether generated latent tokens are actually used by subsequent textual reasoning. We aim to leverage latent visual reasoning as a post-training signal without forcing every response to contain latent tokens. The method design is based on a simple intuition from decoder-only transformers, where later tokens interact with earlier tokens through multi-head attention~\citep{vaswani2017attention}. If the generated latents contribute to the final answer, the text generated after \latentend should attend to the \latent tokens. We use this interaction as an efficient proxy for latent utilization. Without requiring a task-level decision about when latent reasoning should be used, this reward encourages latent tokens to be influential when the model chooses to generate them.

\textbf{Latent attention reward.} 
Let $\mathbf{o}_i = \pi_\theta(q)$ be a response produced by a latent visual reasoning policy $\pi_\theta$ given prompts $q$.
When the response triggers latent mode, it has the format where textual reasoning tokens follow \latentend. Let $T_L$ denote the indices of the \latent tokens, and $T_t$ denote the indices of subsequent text tokens. Let $A_{j,k}$ denote the attention weight from token $j$ to token $k$ in the first decoder layer, averaged over all attention heads, where $k<j$ under causal attention. We define the latent attention reward as the average attention mass assigned by tokens in $T_t$ to tokens in $T_L$, normalized by the number of post-latent text tokens:
\begin{equation}\label{eq:r_a}
R_A^i = \frac{1}{|T_t|} \sum_{j \in T_t}\sum_{k \in T_L} A_{j,k}.
\end{equation}
We compute the attention score at the first decoder layer, where text tokens attend directly to latent tokens before information is further mixed in deeper layers. If a response does not contain generated latents, we set the latent attention reward to zero. Empirically, we do not observe excessive attention on latent tokens (\autoref{fig:attention} right), so no clipping is applied to $R_A$. 

For VLPO~\cite{wang2025monet}, we add the latent attention reward to the standard outcome rewards $r_i$ before calculating group advantage $\hat{A}_{i}$. Specifically, the scalar reward for response $\mathbf{o}_i$ is:
\begin{equation}\label{eq:weights}
r_i =
w_1 R_{\text{acc}}(\mathbf{o}_i)
+
w_2 R_{\text{format}}(\mathbf{o}_i)
+
w_3 R_A^i,
\end{equation}
where $R_{\text{acc}}$ is the correctness reward, $R_{\text{format}}$ is the format reward to make sure the final answer is wrapped in ``\texttt{\textbackslash boxed\{\}}''~\cite{wang2025monet}, and $R_A^i$ is the latent attention reward. The group-normalized advantage is then computed as: $\hat{A}_{i}
=
\frac{
r_i-\operatorname{mean}(\{r_1,r_2,\ldots,r_G\})
}{
\operatorname{std}(\{r_1,r_2,\ldots,r_G\})
}$. The final objective is:
\begin{equation}
    \begin{aligned}
        \mathcal{J}(\theta)
        =
        \mathbb{E}_{q, o\sim \pi_{\text{old}}}
        \frac{1}{G} \sum_{i=1}^G
        \frac{1}{|\mathbf{o}_i|}
        \sum_{t=1}^{|\mathbf{o}_i|}
        \min \Big[
        r_{i,t}(\theta)\hat{A}_{i},
        \operatorname{clip}\big(r_{i,t}(\theta),1-\varepsilon,1+\varepsilon\big)\hat{A}_{i}
        \Big]
        -
        \beta \operatorname{KL}\left(\pi_\theta \| \pi_{\text{ref}}\right).
    \end{aligned}
    \label{eq:objective}
\end{equation}

Following VLPO, the token-level ratio $r_{i,t}(\theta)$ is defined differently for continuous latent tokens and discrete text tokens:
\begin{equation}
r_{i,t}(\theta)=
\begin{cases}
 \exp \left(-\frac{1}{2\sigma^2}\|\mathbf{h}^{\text{old}}_{i,t}-\mathbf{h}^{\theta}_{i,t}\|^2 \right),
& \text{if } t \in T_L, \\[8pt]

\frac{\pi_\theta\left(\mathbf{o}_{i,t} \mid q, \mathbf{o}_{i,<t}\right)}
{\pi_{\theta_{\text{old}}}\left(\mathbf{o}_{i,t} \mid q, \mathbf{o}_{i,<t}\right)},
& \text{if } t \notin T_L.
\end{cases}
\end{equation}
Here, $\mathbf{h}^{\text{old}}_{i,t}$ and $\mathbf{h}^{\theta}_{i,t}$ denote the latent representations produced by the old and current policies, respectively, and $\mathbf{o}_{i,<t}$ denotes the prefix before position $t$. In practice, we use $w_1=0.9$, $w_2=0.1$, and $w_3=1.0$   to ensure that all terms are of comparable magnitude within the interval $[0, 1]$.

This design supports the central goal of leveraging latent visual reasoning ``\textbf{in silence}.'' The attention reward encourages generated latent tokens to interact with subsequent textual reasoning during RL, allowing the model to extract useful visual grounding signals from the latent mode. It places no requirement on the final model to preserve latent generation as a fixed inference-time format.

\section{Experiment Results}\label{sec:experiment_settings}

\subsection{Settings}
We evaluate whether the latent attention reward can improve post-training when added to RL objectives. 
We continue training Monet-SFT and LVR-SFT$^*$ on Thyme-RL with VLPO. Due to computational constraints, the models are trained with a batch size of $4$ for $600$ and $400$ steps, respectively. More training hyperparameters, including learning rate and KL coefficient, are provided in \S\ref{sec:rl-post-training}.

We evaluate the trained models on various multimodal benchmarks, including MathVista-Mini (MathVista)~\citep{lu2023mathvista}, MathVerse-Mini (MathVerse)~\citep{zhang2024mathverse}, MMVP~\citep{zhang2024mmvp}, MMBench-V11-Mini (MMBench)~\citep{liu2024mmbench}, MME-RealWorld-Lite (MME-RW)~\citep{zhang2024mme}, V*~\citep{cheng2025v}, and HRBench4K~\citep{wang2025divide}, comprehensively covering math reasoning, spatial reasoning, and fine-grained visual perception tasks. We use VLMEvalKit~\citep{duan2024vlmevalkit} with Gemini-2.5-Pro~\citep{comanici2025gemini} / Gemini-3.1-Pro~\citep{gemini31pro_modelcard_2026} as a judge following prior work~\citep{wang2025monet}.

\subsection{Main Results}\label{sec:result}

\begin{wrapfigure}{r}{0.6\textwidth}
\vspace{-20pt}
    \centering
    \includegraphics[width=\linewidth]{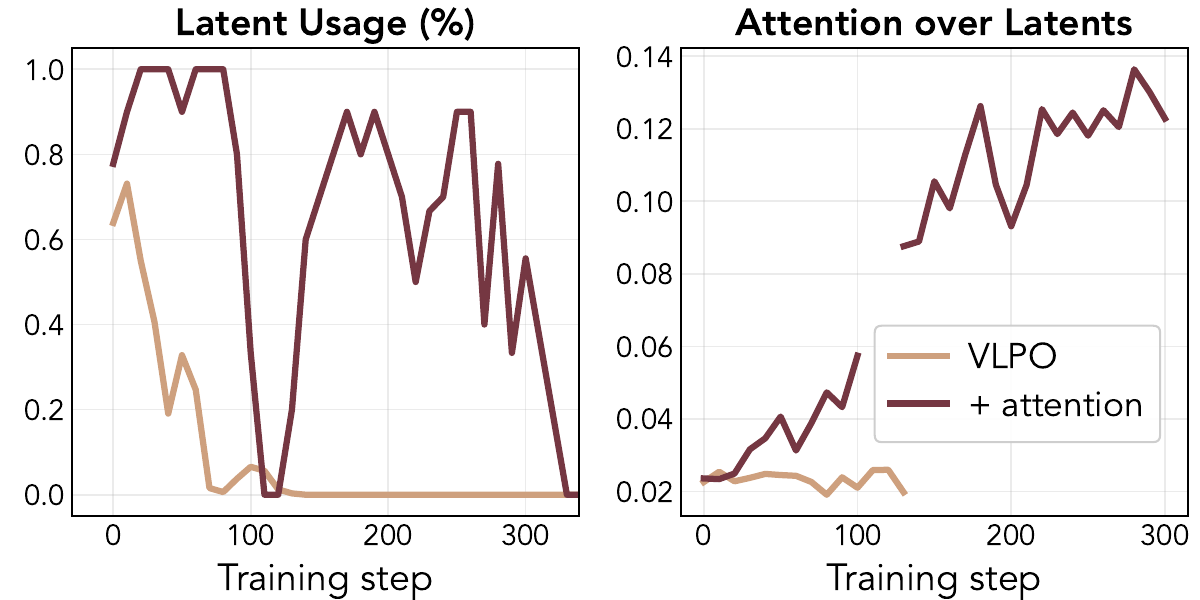}
    \caption{
    \label{fig:attention} Attention reward encourages visual latents more attended to by subsequent textual reasoning. The gap in attention corresponds to the lack of latents around step $100$.}
\vspace{-10pt}
\end{wrapfigure}

We first analyze how the attention reward changes latent behavior during RL. The right panel of \autoref{fig:attention} shows the attention from subsequent text tokens to generated latent tokens for samples that trigger the latent mode, where adding $R_A$ substantially increases this attention mass. At the end of training, this interaction is approximately $7\times$ higher than vanilla VLPO. This indicates that the reward successfully encourages generated latent tokens to participate in textual reasoning. The left panel of \autoref{fig:attention} shows that the model also triggers the latent mode more frequently under the attention reward, allowing extended time to exploit potential knowledge from visual latents. 
As training progresses, the model gradually prunes low-value latent usages and retains only highly interactive latent responses before eventually internalizing the learned behavior into pure-text generation. Eventually, the latent usage ratio converges to zero.

\begin{table}[t]
    \centering
    \caption{\textbf{Performance on perception and reasoning benchmarks}. Results with ``$\dagger$'' are reported by other papers, otherwise evaluated by us. \textbf{Bold} entries denote that our method yields the best performance among the model family. $\color{PineGreen}{\uparrow}$ denote our improvement compared to the SFT base model.}
    \label{tab:monet}
    \small
    \setlength{\tabcolsep}{3.5pt}
    \resizebox{\textwidth}{!}{
    \begin{tabular}{@{}lLLLLLLLLLLL@{}}
        \toprule
        \multirow{2}{*}{\textbf{Model}} 
        & \textbf{MathVista} 
        & \textbf{MathVerse}
        & \textbf{MMVP} 
        & \textbf{MMBench} 
        & \textbf{MME-RW}
        & \multicolumn{3}{c}{\textbf{V*}} 
        & \multicolumn{3}{c}{\textbf{HRBench4K}} \\
        \cmidrule(lr){2-2}
        \cmidrule(lr){3-3}
        \cmidrule(lr){4-4}
        \cmidrule(lr){5-5}
        \cmidrule(lr){6-6}
        \cmidrule(lr){7-9}
        \cmidrule(lr){10-12}
        & \text{Overall} & \text{Overall} & \text{Overall} & \text{Overall} 
        & \text{Overall}
        & \text{Overall} & \text{Attr.} & \text{Spat.} 
        & \text{Overall} & \text{FSP} & \text{FCP} \\
        \midrule
        GPT-4o
        & - & - & - & - 
        & 52.0^\dagger
        & 67.5^\dagger & 72.2^\dagger & 60.5^\dagger
        & 59.0^\dagger & 70.0^\dagger & 48.0^\dagger \\
        
        Qwen2.5-VL-7B
        & 68.2 & 48.0 & 76.7 & 84.3
        & 45.8
        & 76.4 & 77.4 & 75.0
        & 69.3 & 85.3 & 53.3 \\
        
        Deepeyes
        & - & - & - & -
        & 54.3^\dagger
        & 83.3^\dagger & 84.4^\dagger & 81.6^\dagger
        & 71.3^\dagger & 83.8^\dagger & 58.8^\dagger \\
        
        SkiLa
        & - & - & 75.3^\dagger & 83.3^\dagger
        & 56.6^\dagger
        & 84.3^\dagger & - & -
        & 72.0^\dagger & - & - \\
        
        CoVT
        & - & - & - & -
        & 63.3^\dagger
        & 78.5^\dagger & - & -
        & 72.5^\dagger & - & - \\
        
        \midrule
        LVR-SFT w/o latents
        & - & - & 72.0^\dagger & -
        & -
        & 79.1^\dagger & 82.6^\dagger & 73.7^\dagger
        & - & - & - \\ 

        LVR-SFT$^*$ 
        & 63.0 & 36.8 & 75.7
        & 77.8
        & 50.4
        & 82.7 & 85.2 & 78.9
        & 70.5 & 80.8 & 60.3 \\

        LVR-SFT$^*$ + VLPO
        & 61.5 & 37.3 & 76.7
        & 83.3
        & 50.8
        & 81.7 & 82.6 & 80.3
        & 71.5 & 83.3 & 59.8 \\

        \rowcolor[HTML]{F7CCD4}
        LVR-SFT$^*$ + VLPO + Attn (Ours)
        & \mathbf{63.7} & 36.3 & \mathbf{77.7}
        & 82.4
        & \mathbf{50.8}
        & \mathbf{85.9} & \mathbf{86.1} & \mathbf{85.5}
        & \mathbf{71.9} & \mathbf{84.5} & 59.3 \\
        
        \quad $\Delta$ vs. SFT
        & {\color{PineGreen} \uparrow0.7} & {\color{purple} \downarrow0.5} & {\color{PineGreen} \uparrow2.0} 
        & {\color{PineGreen} \uparrow4.6} 
        & {\color{PineGreen} \uparrow0.4} 
        & {\color{PineGreen} \uparrow3.2} & {\color{PineGreen} \uparrow0.9} & {\color{PineGreen} \uparrow6.6}
        & {\color{PineGreen} \uparrow1.4} & {\color{PineGreen} \uparrow3.7} & {\color{purple} \downarrow1.0} \\
        
        \midrule
        Monet-SFT w/o latents
        & - & - & - & -
        & 51.3^\dagger
        & 81.7^\dagger & 83.5^\dagger & 79.0^\dagger
        & 68.4^\dagger & 78.3^\dagger & 58.5^\dagger \\ 
        
        Monet-SFT w/o latents + GRPO
        & - & - & - & -
        & 52.4^\dagger
        & 78.5^\dagger & 78.3^\dagger & 79.0^\dagger
        & 70.0^\dagger & 83.3^\dagger & 56.8^\dagger \\ 
        
        Monet-SFT
        & 60.5 & 35.1 & 73.0 & 74.1
        & 52.7
        & 80.1 & 82.6 & 76.3
        & 68.5 & 79.5 & 56.5 \\
        
        Monet-SFT + VLPO
        & 59.9 & 36.5 & 65.7 & 75.9
        & 55.0
        & 81.2 & 81.7 & 80.3
        & 70.4 & 81.3 & 59.3 \\
        
        \rowcolor[HTML]{F7CCD4}
        Monet-SFT + VLPO + Attn (Ours)
        & \mathbf{67.9} & \mathbf{39.9} & \mathbf{76.0} & \mathbf{77.8}
        & 52.8
        & \mathbf{85.3} & \mathbf{85.2} & \mathbf{85.5}
        & \mathbf{71.1} & \mathbf{85.5} & 56.8 \\
        
        \quad $\Delta$ vs. SFT
        & {\color{PineGreen} \uparrow7.4} & {\color{PineGreen} \uparrow4.8} & {\color{PineGreen} \uparrow3.0} & {\color{PineGreen} \uparrow3.7}
        & {\color{PineGreen} \uparrow0.1}
        & {\color{PineGreen} \uparrow5.2} & {\color{PineGreen} \uparrow2.6} & {\color{PineGreen} \uparrow9.2}
        & {\color{PineGreen} \uparrow2.6} & {\color{PineGreen} \uparrow6.0} & {\color{PineGreen} \uparrow0.3} \\
        
        \bottomrule
    \end{tabular}
    }
\end{table}

\autoref{tab:monet} reports the main benchmark results. For both Monet and LVR, adding the attention reward improves the SFT model on most metrics. The largest gains are on visually grounded benchmarks, especially V* spatial reasoning (+$9.2$, +$6.6$) and HRBench4K-FSP (+$6.0$, +$3.7$). The reward also improves the performance of Monet on MathVista by $7.4$. These results suggest that the attention reward helps the base model better exploit visual evidence during post-training. 
Our method shows even larger gains on MathVista (+$8.0$) and MMVP (+$10.3$) than VLPO for the Monet baseline, demonstrating the effectiveness of the attention reward in improving visual reasoning capabilities.

Overall, the results support the claim that latent generation alone is not a reliable indicator of better performance, and standard RL may suppress latent tokens when they are not consistently necessary. The attention reward encourages latent tokens to influence textual reasoning more softly than forcing the latent generation for every sample. While the latent mode is still diminished at the end of training, the RL training helps the model shape more accurate visual-related textual reasoning in silence.

\begin{figure}[t]
    \centering
    \begin{minipage}{0.61\textwidth}
        \centering
        \captionof{table}{\textbf{Ablation on latent type in attention reward}. Using random latents in VLPO degrades performance compared with real latents. Both models are trained for $400$ steps.}
        \label{tab:ablation_rand_table}
        \small
        \setlength{\tabcolsep}{3.5pt}
        \resizebox{\linewidth}{!}{
        \begin{tabular}{@{}lLLLLL@{}}
            \toprule
            \multirow{1}{*}{\textbf{Reward Type}} 
            & \textbf{MathVista} 
            & \textbf{MMVP} 
            & \textbf{MMBench} 
            & \textbf{V*}
            & \textbf{HRBench4K} \\
            \midrule
            Rand. Lat. + $R_A$
            & 60.9 & 73.0 & 75.0 & 80.6 & 65.6 \\
            Real Lat. + $R_A$
            & \mathbf{62.9} & \mathbf{74.3} & \mathbf{80.6} & \mathbf{85.3} & 67.9 \\
            \bottomrule
        \end{tabular}
        }
    \end{minipage}
    \hfill
    \begin{minipage}{0.36\textwidth}
        \centering
        \includegraphics[width=\linewidth, height=4cm, keepaspectratio]{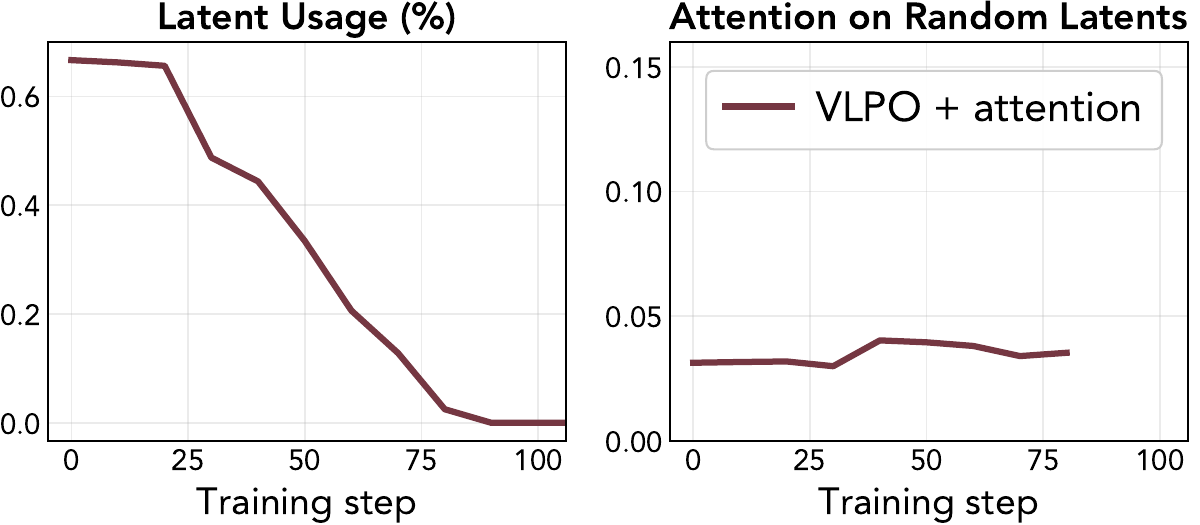}
        \vspace{-10pt}
        \caption{\textbf{Attention on random latents during VLPO training}.}
        \label{fig:ablation_rand_figure}
    \end{minipage}
\vspace{-12pt}
\end{figure}

\subsection{Analysis}
\textbf{Necessity of latent tokens for attention reward.}
To verify that the benefit of $R_A$ comes from meaningful visual latents, we conduct an ablation that replaces real latent embeddings with random values during RL rollout. Following \S\ref{sec:latent_dependence}, each generated latent embedding is replaced with Gaussian noise before subsequent token generation. We then use VLPO + $R_A$ to post-train Monet-SFT. As shown in \autoref{fig:ablation_rand_figure}, the latent mode collapses within the first $100$ steps. Meanwhile, the attention assigned to random latents remains low, indicating that the model does not exploit $R_A$ by attending to uninformative latent tokens. This addresses a potential reward-hacking concern, where the reward could be satisfied by arbitrary latent attention.

We also report the performance comparison of using random latents vs. real latents in Table \ref{tab:ablation_rand_table}. The results show that training with real latents consistently outperforms training with random latents across all five benchmarks, with an average gain of +$3.2$ points. The attention reward gradually improves visual grounding during post-training. The effect cannot be 
equally achieved with random latents. Through this ablation study, we confirm that the effectiveness of the attention reward comes from interaction with meaningful visual latent representations.

\noindent \textbf{Attribution analysis for the attention reward.}\label{sec:app-causal}
The proposed attention reward uses attention to latent tokens as a signal to indicate the use of generated latents. We use two complementary gradient-based attribution analyses to examine whether latent representations with higher attention values also influence the probability of the final answer more strongly.

Specifically, we save checkpoints at training steps $0$, $30$, and $60$, before latent generation disappears under VLPO. We evaluate responses containing latent tokens on $300$ held-out questions sampled from Thyme-RL, MMVP, and V*. For each response, we define the target as the length-normalized log-probability of the final answer span. We then compute attribution from the latent tokens using Gradient x Input (GxI,~\cite{shrikumar2017learning}) and Integrated Gradients (IG,~\cite{sundararajan2017axiomatic}). Larger values indicate that the answer likelihood is more sensitive to the generated latent representations. \autoref{tab:attribution-scores} shows that as training progresses, our model generates latents that are more important for producing final answers than the original SFT model (Step $0$) and the VLPO model. This trend aligns with the attention change in \autoref{fig:attention} for both VLPO and our model.

\begin{table}[H]
    \centering
    \vspace{-6pt}
    \footnotesize
    \setlength{\tabcolsep}{3pt}
    \begin{minipage}[t]{0.44\textwidth}
        \centering
        \caption{Attribution scores across training steps. GxI are scaled by $10^{3}$ and IG by $10^{2}$.}
        \label{tab:attribution-scores}
        \resizebox{\linewidth}{!}{%
        \begin{tabular}{@{}lLLLLLL@{}}
            \toprule
            \multirow{2}{*}{\textbf{Model}}
            & \multicolumn{2}{c}{\textbf{Step $0$}}
            & \multicolumn{2}{c}{\textbf{Step $30$}}
            & \multicolumn{2}{c}{\textbf{Step $60$}} \\
            \cmidrule(lr){2-3}
            \cmidrule(lr){4-5}
            \cmidrule(lr){6-7}
            & \text{GxI} & \text{IG} & \text{GxI} & \text{IG} & \text{GxI} & \text{IG} \\
            \midrule
            Ours
            & 0.62 & 3.63 & \mathbf{-0.35} & \mathbf{7.57} & \mathbf{10.56} & \mathbf{20.95} \\
            VLPO
            & 0.62 & 3.63 & \mathbf{-}7.19 & 7.22 & 0.45 & 3.31 \\
            \bottomrule
        \end{tabular}%
        }
    \end{minipage}%
    \hfill
    \begin{minipage}[t]{0.51\textwidth}
        \centering
        \caption{Correlation between attention scores and attribution scores.}
        \label{tab:correlation-scores}
        \resizebox{\linewidth}{!}{%
        \begin{tabular}{@{}lLLLLLL@{}}
            \toprule
            \multirow{2}{*}{\textbf{Model}}
            & \multicolumn{2}{c}{\textbf{Step $0$}}
            & \multicolumn{2}{c}{\textbf{Step $30$}}
            & \multicolumn{2}{c}{\textbf{Step $60$}} \\
            \cmidrule(lr){2-3}
            \cmidrule(lr){4-5}
            \cmidrule(lr){6-7}
            & \text{GxI} & \text{IG} & \text{GxI} & \text{IG} & \text{GxI} & \text{IG} \\
            \midrule
            Ours
            & -0.073 & 0.093 & \mathbf{0.160} & \mathbf{-0.050} & \mathbf{0.247} & \mathbf{0.301} \\
            VLPO
            & -0.073 & 0.093 & 0.032 & -0.082 & 0.062 & -0.026 \\
            \bottomrule
        \end{tabular}%
        }
    \end{minipage}
    \vspace{-8pt}

\end{table}

We next compute the correlation between attention scores on latents and the attribution scores above. The results in \autoref{tab:correlation-scores} indicate that the correlation between attention scores and attribution scores is increasingly positive for our method. This supports the idea that the proposed reward drives training to identify latent representations that help derive the correct answers.

\noindent \textbf{Source of gain for latent-free final models.}\label{sec:app-final}
We also probe the final model's behavior to understand its better performance. Because the final model emits no latents, we measure what its pure-text inference actually uses, applying the same gradient-attribution analysis as above to image tokens with Gradient $\times$ Input and Integrated Gradients. 
An image contributes hundreds of tokens, but most are irrelevant to any given query. We expect the model to focus on discriminative regions to derive the answer.
Accordingly, we report the normalized entropy of per-image-token attribution, where lower values mean a smaller subset of the image tokens influence the output more significantly. 

\begin{table}[H]
\centering
\begin{minipage}[t]{0.58\linewidth}
    \centering
    \vspace{-6pt}
    \captionof{table}{Entropy of per-image-token gradient attribution before (Step $0$) and after training (Step $600$).}
    \label{tab:attribution-entropy}
    \small
    \setlength{\tabcolsep}{5pt}
    \begin{tabular}{@{}lCCCC@{}}
        \toprule
        \multirow{2}{*}{\textbf{Model}} & \multicolumn{2}{c}{\textbf{Step $0$}} & \multicolumn{2}{c}{\textbf{Step $600$}} \\
            \cmidrule(lr){2-3}
            \cmidrule(lr){4-5}
        & \text{GxI} & \text{IG} & \text{GxI} & \text{IG} \\
        \midrule
        Ours & 0.859 & 0.898 & \mathbf{0.849} & \mathbf{0.880} \\
        VLPO & 0.859 & 0.898 & 0.854 & 0.891 \\
        \bottomrule
    \end{tabular}
\end{minipage}%
\hfill
\begin{minipage}[t]{0.4\linewidth}
    \centering
    \vspace{-6pt}
    \captionof{table}{Alignment of attribution with ground-truth important regions (AUC).}
    \label{tab:auc-comparison}
    \small
    \setlength{\tabcolsep}{5pt}
    \begin{tabular}{@{}lC@{}}
        \toprule
        & \textbf{AUC} \\
        \midrule
        SFT & 0.654 \\
        VLPO & 0.639 \\
        Ours & \mathbf{0.669} \\
        \bottomrule
    \end{tabular}
\end{minipage}
\vspace{-8pt}
\end{table}

As shown in \autoref{tab:attribution-entropy}, image token attribution in our model is more concentrated (lower entropy) than both the final VLPO model and the initial SFT model (Step $0$), and the reduction from initialization is roughly twice VLPO's ($-0.010$/$-0.018$ vs $-0.005$/$-0.007$). Crucially, total attention mass over image tokens is matched between the two models ($\approx 0.025$).
Therefore, the two models differ mainly in how selectively they use image content to derive the answers.
We hypothesize that the attention reward enables the model to focus on discriminative image regions during text reasoning.

To test whether attribution is focusing on the relevant part of the image rather than being randomly concentrated, we evaluate it against the ground-truth important regions annotated in 200 held-out samples from Visual-CoT ~\cite{shao2024visual}. We treat the per-image-token attribution as a score for the binary label ``this token lies inside the annotated region'' and report scale-invariant ROC-AUC, \textit{i.e.}, the probability that a randomly drawn in-region token receives higher attribution than a randomly drawn out-of-region token, as the metric for alignment of attribution with relevant image regions.

\autoref{tab:auc-comparison} shows that our model ranks the annotated regions highest, while VLPO yields worse alignment than the shared SFT initialization. This is consistent with the qualitative pattern in \S\ref{sec:app-examples} in the main paper, where Monet-VLPO produces fluent reasoning that drifts from the image. Together, these results show that our final model reads the same amount of the image but relies more selectively on the regions that matter, a property that is beyond latent generation and plausibly explains the retained gain after latent disappears.

\noindent \textbf{Why is latent token generation discarded during RL training?}
During training, latent responses are no more accurate than pure-text responses (\autoref{tab:rollout_stats}), and since the correctness of the answer still dominates the objective (\autoref{eq:weights}). Thus, latent generation eventually diminishes.

\begin{figure}[t]
    \centering
    \includegraphics[width=\linewidth]{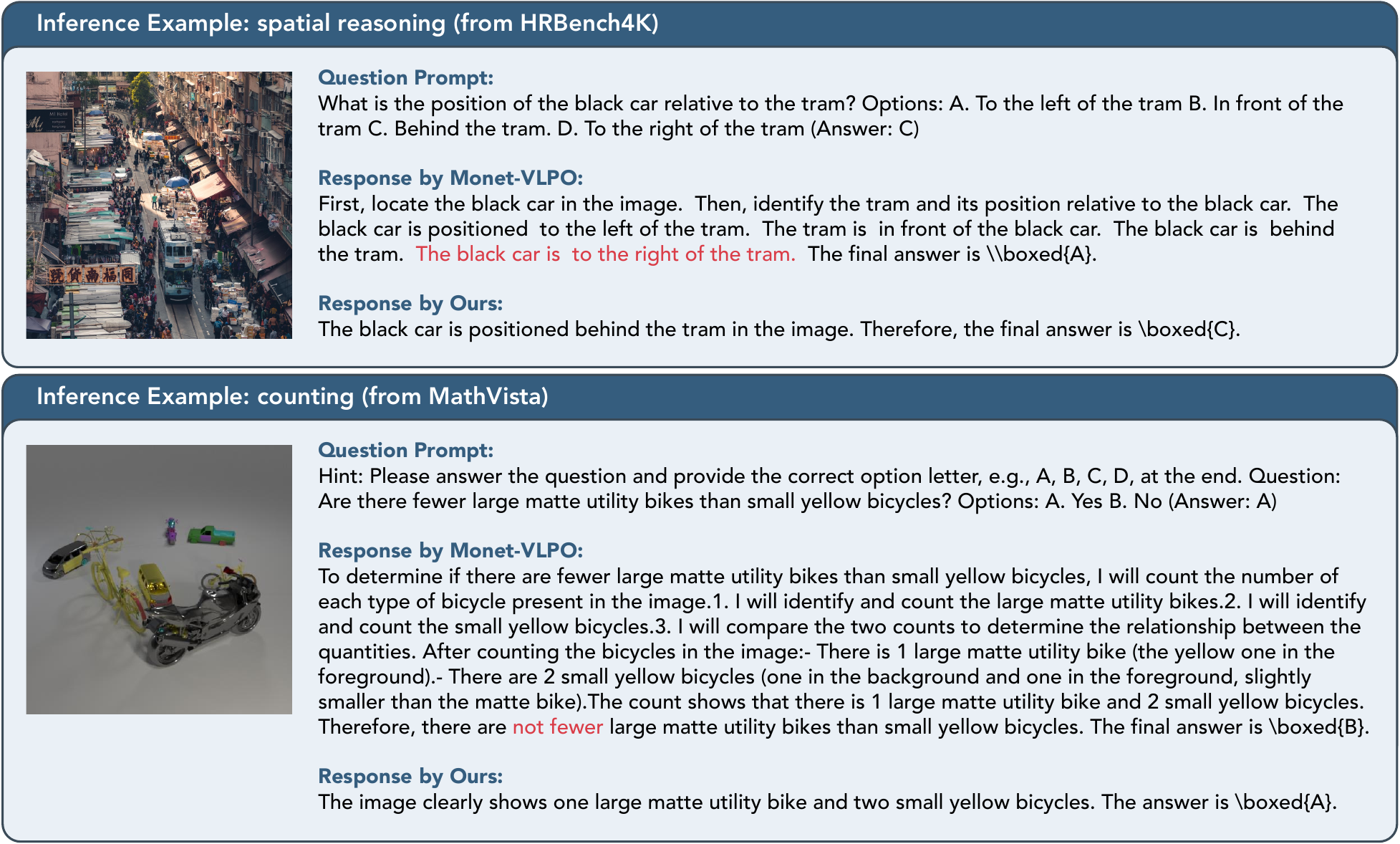}
    \caption{Qualitative comparison on reasoning examples. Monet-VLPO reasons longer but drifts from the image evidence. Our model responds concisely and stays visually grounded.}
    \label{fig:examples}
\end{figure}

\subsection{Case Study}\label{sec:qualitative}
The case studies in \autoref{fig:examples} illustrate how our method improves visual grounding during inference. In the HRBench4K spatial reasoning example, Monet-VLPO produces a long reasoning chain but fails to maintain a consistent spatial relation, leading to a wrong answer. Our model directly identifies the relation between the black car and the tram and predicts the correct answer. In the counting example, Monet-VLPO decomposes the problem into several counting steps, yet its final comparison contradicts the visual evidence. Our model gives a shorter response and correctly counts one large matte utility bike and two small yellow bicycles. These examples suggest that stronger visual grounding can reduce reasoning drift and lead to more concise answers aligned with the image.
\label{sec:shorter_response}
The conciseness may also relate to the design of the attention reward. In transformers, attention from later text tokens to a fixed early prefix can naturally decrease as the context grows. Longer continuations therefore reduce average attention to latent tokens. This formulation may encourage the model to generate shorter post-latent textual reasoning while maintaining interaction with visual latents. As a result, our model has improved response time as reported in \S\ref{sec:app-speed} (more analyses in \S\ref{sec:app-length}).
Overall, these examples support the central finding of our paper: latent visual reasoning does not need to persist as an explicit inference-time format to be useful. Even when the final response is expressed mostly in text, the training signal from latent-token interaction can help the model shape more accurate visual grounding behavior. More inference examples are listed in \S\ref{sec:app-examples}.

\section{Conclusion}
This paper studies how latent visual reasoning behaves during inference and RL post-training. We show that generated latent tokens are often used only weakly to obtain the correct answer, and that RL drives models to skip the latent mode for most questions. 
We therefore shift the focus from latent necessity to latent utilization. We propose a latent attention reward that encourages generated latents to influence subsequent textual reasoning during RL.
Experiments show that this reward improves performance on visual reasoning benchmarks, especially those requiring stronger visual grounding. Even when latent generation decreases after post-training, the learned policy can still benefit from the training signal provided by latent tokens in silence. 
We hope this perspective encourages a broader view of latent reasoning as a mechanism for improving visual grounding during learning, even when its benefits are ultimately expressed through ordinary textual responses.

\begin{ack}
This research is supported in part by grants from the National Science Foundation (OAC-2118240, HDR Institute: Imageomics). We also thank Prof. Ziang Xiao from Johns Hopkins University for his support.
\end{ack}

\bibliography{neurips_2026_references}
\bibliographystyle{plainnat}

\appendix
\clearpage
\section*{Appendix}

The appendix is organized as follows. 
\begin{itemize}[leftmargin=18pt,nosep]
    \item In \S\ref{sec:app-impact}, we discuss the societal impact of this research.
    \item In \S\ref{sec:limitation}, we discuss the current limitations of this research as well as promising future directions.
    \item In \S\ref{sec:app-lvr-sft}, we describe the construction of LVR-SFT* with random latent dropping. 
    \item In \S\ref{sec:rl-post-training}, we provide implementation details for RL post-training, including the lightweight training setup and hyperparameters used for the attention reward.
    \item In \S\ref{sec:app-necessity}, we present the full formulation of the diagnostic latent necessity reward.
    app-length
    \item In \S\ref{sec:app-empirical}, we report additional empirical results, including the comparison between zero-latent and standard latent rollouts and the inference speed analysis. We also show how controlling response length itself is not necessarily beneficial, while combining with our method can partly mitigate its effect.
    \item In \S\ref{sec:app-examples}, we provide additional qualitative examples across benchmarks that further illustrate the behavior of our method.
\end{itemize}

\section{Broader Impacts}\label{sec:app-impact}
This work improves visual grounding in multimodal reasoning models by better leveraging latent visual reasoning during post-training. Stronger visual grounding can benefit applications that require reliable interpretation of visual evidence. By encouraging models to make better use of intermediate visual representations, our method may reduce unsupported textual reasoning and improve alignment between generated answers and image content.

At the same time, our method is built on existing multimodal models and inherits their general risks, including potential hallucination and dataset bias. Improved visual reasoning could also increase user trust in model outputs, so deployment in high-stakes settings should still involve careful validation and appropriate human oversight. We do not anticipate misuse risks beyond those associated with the base models, although downstream applications should follow the same safety and privacy considerations required for multimodal systems.

\section{Limitations and Future Work}\label{sec:limitation}
Our study focuses on two representative latent visual reasoning frameworks, Monet and LVR. These models cover common designs for injecting continuous visual latents into multimodal reasoning, yet future methods may use different latent construction mechanisms. It would be valuable to examine whether the observed RL dynamics and the benefits of the attention reward hold across a broader range of latent reasoning architectures.

Our attention reward uses attention mass as a proxy for latent utilization. This provides an efficient training signal. Future work could combine attention-based rewards with stronger counterfactual measurements of latent contribution. In addition, our experiments are conducted under limited computational budgets, so larger-scale post-training and broader hyperparameter sweeps may further clarify when latent reasoning is most beneficial.

Finally, our results suggest that latent reasoning can improve learning even when latent generation becomes rare at inference time. This raises new questions about how models improve latent-guided behavior and when explicit latent generation should be preserved. Future work may explore adaptive trigger policies, example-dependent latent budgets, and reward designs that better balance latent utilization, response quality, and inference efficiency.

\section{Details of LVR-SFT\texorpdfstring{$^*$}{*} Training}\label{sec:app-lvr-sft}
As stated in \S\ref{sec:latent-behavior}, the training samples used in the officially-released LVR-SFT model strictly follow the latent-generation format~\cite{li2025latent}. Therefore, the SFT model is not flexible in choosing whether to generate latents or not. Therefore, we perform an additional SFT stage with the latent target randomly disabled at a probability of $0.4$.

Here we report the training details of the model. It is trained using $2$ H100 GPUs, each of 80GB VRAM for $200$ steps, with data packing of $4$ instances per batch and longest single instance of $4096$ tokens. We keep the vision head frozen, visual token range of min=$128$ and max=$5120$. We set the learning rate=$1\mathrm{e}{-5}$, weight decay=$0.1$, warmup ratio=$0.03$, and LVR $\lambda$=$0.1$. 
Part of the performance comparison between the official LVR-SFT and our LVR-SFT$^*$ is listed in \autoref{tab:monet}. We also report more comparisons on LVR's adopted benchmarks in \autoref{tab:appendix_lvrsft2}. The results show that LVR-SFT$^*$ has a comparable performance to the original LVR-SFT~\citep{li2025latent}.
\begin{table}[H]
    \begin{center}
    \caption{Performance comparison between LVR-SFT* compared and original LVR-SFT~\citep{li2025latent} (Continued). During inference, both latent models use $8$ latent tokens, and the temperature is set to $0$.}
    \vspace{2pt}
    \label{tab:appendix_lvrsft2}
        \resizebox{\linewidth}{!}{
        \begin{tabular}{@{}lLLLLLLLLLLL@{}}
            \toprule
            Method & V^* & V^*_{D.A.} & V^*_{R.P.} & \text{MMVP} & \text{BLINK(overall)} & \text{Counting} & \text{IQ-Test} & \text{JigSaw} & \text{Relative Depth} & \text{Relative Reflect} & \text{Spatial Relation}   \\
            \midrule
            Qwen2.5-VL-7B-Instruct                                   & 79.6 & 81.7 & 76.3 & 67.0 & 58.3 & 67.5 & 27.3 & 51.3 & 79.0 & 41.8 & \mathbf{88.1} \\
            LVR-SFT               & 81.7 & 84.4 & 77.6 & \mathbf{72.0} & 58.8 & \mathbf{70.8} & 29.3 & \mathbf{52.7} & 77.4 & 41.8 & 86.0 \\
            LVR-SFT*     & \mathbf{82.2} & \mathbf{85.2} & \mathbf{77.6} & 71.3 & \mathbf{59.1} & 68.3 & \mathbf{30.7} & 52.0 & \mathbf{81.5} & \mathbf{44.0} & 83.2 \\
            \bottomrule
        \end{tabular}
        }
    \end{center}
\end{table}

\section{Details of RL Post-Training}\label{sec:rl-post-training}

\textbf{Light training in \S\ref{sec:latent-behavior}.}
The official Monet-VLPO~\citep{wang2025monet} is trained on $3.2$K samples from Thyme-RL~\citep{zhang2025thyme} for 1 epoch using a batch size of $64$. Due to limited GPU availability (at most $2$ GPUs for $8$ hours per run), we train with a batch size of $4$. This does not affect the size of the rollout group per sample.

\textbf{Training with attention reward.}
We use the officially released model and code from Monet~\citep{wang2025monet} and train the Monet-SFT-stage3 model and LVR-SFT$^*$ model on Thyme-RL dataset using verl~\citep{sheng2025hybridflow}, with the following hyperparameters:
\begin{table}[H]
    \centering
    \caption{\textbf{Hyperparameters for RL.}}
    \label{tab:rl_params_training}
    \small
    \vspace{2pt}
    \begin{tabular}{lclc}
        \toprule 
        Hyperparameter & Value & Hyperparameter & Value \\
        \midrule 
        learning rate & $1\mathrm{e}{-6}$ & batch size & $4$ \\
        weight decay & $0.01$ & rollout size & $8$ \\
        temperature & $0.5$ & max response length & $4096$ \\
        VLPO $\sigma$ & $10.0$ & latent size & $10$ \\
        max pixels per img & $2000 \times 28 \times 28$ & accuracy threshold & $0.6$ \\
        training steps & $600$ (Monet) $400$ (LVR) & accelerator & $2$ NVIDIA H100-80GB \\
        number of visual latents & $10$ & data type & bfloat16 \\
        \bottomrule
    \end{tabular}
\end{table}
\label{app-k-latents}

The reward itself introduces no additional trainable module. It only extracts attention from the first decoder layer during the log-probability computation of the old policy. The practical runtime and peak vram for our method is 82.2 H100 hours and 78.90 GB, compared to VLPO's 82.0 H100 hours and 76.11 GB. The small runtime and memory increases indicate that computing the reward will not be a scaling bottleneck. This supports computational compatibility with larger models.

Due to the substantial compute and API cost of latent-space RL, fully repeating both VLPO and our method across several seeds and architectures exceeds our available budget. The current one-run protocol follows the previous works, where LVR and Monet don’t report standard deviation either. Moreover, our conclusion does not depend on a single evaluation set. The models are evaluated on seven benchmarks, and an overall average improvement is observed across two backbones.

\section{Details of Latent Necessity Reward}\label{sec:app-necessity}

 Let $\mathbf{o}_i = \pi_\theta(q)$ be a response containing latent tokens produced by a latent visual reasoning policy $\pi_\theta$ given prompts $q$. $\mathbf{o}_i$ has the format $\mathbf{o}_i = \verb|<text1>|\circ\verb|<latent_start>|\circ\verb|<latents>|\circ\verb|<latent_end>|\circ\verb|<text2>|$ where $\verb|<latents>|$ denotes latent visual tokens and $\verb|<text1>|$, $\verb|<text2>|$ denotes text tokens.  We let $\pi_\theta$ continue to generate another response given the same prefix before $\verb|<latents>|$: $\mathbf{o}_j = \pi_\theta(q\circ\verb|<text1>|\circ\verb|<latent_start>|\circ\verb|<latent_end>|$. $\mathbf{o}_j$ is different from spontaneously generated textual responses in that a pure-text response with the same prefix is highly likely to still complete as $\mathbf{o}_i$. The necessity reward $R_N$ for $o_i$ is then defined as:
\begin{equation}
R_N(o_i) = 
\begin{cases}
n, & r_i=1,\ r_j=0 \\
-\frac{n}{2}, & r_i=1,\ r_j=1 \\
-\frac{n}{2}, & r_i=0,\ r_j=1 \\
0, & r_i=0,\ r_j=0 \\
\end{cases}
\label{eq:reward_0}
\end{equation}
where $r_i$, $r_j$ denote the accuracies of $o_i$, $o_j$. Intuitively, $R_N$ is positive only if a regeneration without \verb|<latents>| has lower accuracy than the original latent response. We set $N$ = 1 and train Monet-SFT model using VLPO + $R_N$ for 600 steps. As \autoref{tab:app-necessity-perf} shows, this model performs on par with the SFT base model, suggesting that the necessity reward alone is insufficient to incentivize training process to utilize visual latents.

\begin{table}[t]
    \centering
    \caption{\textbf{Performance of Monet-SFT + $R_N$ on perception and reasoning benchmarks}. \textbf{Bold} entries denote the best performance.}
    \label{tab:app-necessity-perf}
    \vspace{2pt}
    \small
    \setlength{\tabcolsep}{3.5pt}
    \resizebox{\textwidth}{!}{
    \begin{tabular}{@{}lLLLLLLLLLL@{}}
        \toprule
        \multirow{2}{*}{\textbf{Model}} 
        & \textbf{MathVista} 
        & \textbf{MMVP} 
        & \textbf{MMBench} 
        & \textbf{MME-RW}
        & \multicolumn{3}{c}{\textbf{V*}} 
        & \multicolumn{3}{c}{\textbf{HRBench4K}} \\
        \cmidrule(lr){2-2}
        \cmidrule(lr){3-3}
        \cmidrule(lr){4-4}
        \cmidrule(lr){5-5}
        \cmidrule(lr){6-8}
        \cmidrule(lr){9-11}
        & \text{Overall} & \text{Overall} & \text{Overall} 
        & \text{Overall}
        & \text{Overall} & \text{Attr.} & \text{Spat.} 
        & \text{Overall} & \text{FSP} & \text{FCP} \\
        \midrule
        Monet-SFT
        & 60.5 & \mathbf{73.0} & 74.1
        & \mathbf{52.7}
        & \mathbf{80.1} & \mathbf{82.6} & \mathbf{76.3}
        & 68.5 & 79.5 & 56.5 \\

        Monet-SFT + VLPO + $R_N$
        & \mathbf{71.3} & 65.0 & \mathbf{76.9}
        & 51.4
        & \mathbf{80.1} & \mathbf{82.6} & \mathbf{76.3}
        & \mathbf{70.5} & \mathbf{84.3} & \mathbf{56.8} \\

        \bottomrule
    \end{tabular}
    }
\end{table}

\section{More Empirical Results}\label{sec:app-empirical}

\subsection{Necessity of Generated Latents in RL}
As shown in \autoref{tab:monet} and the original Monet paper~\cite{wang2025monet}, using latent visual tokens in SFT and later in RL post-training yields better performance than vanilla training without latents. We further investigate the importance of latent visual tokens solely at the RL stage. As the attention reward cannot be applied when there are no latent visual tokens, we conduct this experiment with VLPO. We discard the generated latent tokens when the model triggers the latent mode. The model is trained for $400$ steps and compared with training with 10 latent tokens (default setting for Monet-VLPO). As reported in \autoref{tab:1modeonly_monet}, training with $0$ latent tokens when triggering latent mode is generally worse than training with $10$ latent tokens, suggesting that such a shortcut has inferior potential.
\begin{table}[H]
    \centering
    \caption{Performance of the Monet model trained with 0 latent tokens is worse than that trained with 10 latent tokens.}
    \label{tab:1modeonly_monet}
    \small
    \vspace{2pt}
    \begin{tabular}{@{}lL L L LLL LLL@{}}
        \toprule
        \multirow{2}{*}{\textbf{Model}} 
        & \multicolumn{1}{c}{\textbf{MathVista}} 
        & \multicolumn{1}{c}{\textbf{MMVP}} 
        & \multicolumn{3}{c}{\textbf{V*}} 
        & \multicolumn{3}{c}{\textbf{HRBench4K}} \\
        \cmidrule(lr){2-2}\cmidrule(lr){3-3}\cmidrule(lr){4-6}\cmidrule(lr){7-9}
        & \text{Overall} & \text{Overall}
        & \text{Overall} & \text{Attribute} & \text{Spatial} 
        & \text{Overall} & \text{FSP} & \text{FCP}  \\
        \midrule 

        Monet-7B (SFT)
        & 60.5 & 73.0
        & 80.1 & 82.6 & 76.3
        & 68.0 & 79.5 & 56.5  \\

        \midrule 

        VLPO w/ 0 latents
        & 64.0 & 72.7  
        & 76.4 & 80.0 & 71.1 
        & 71.6 & 85.0 & 58.3 \\

        VLPO w/ 10 latents
        & 65.8 & 73.0 
        & 79.1 & 79.1 & 78.9
        & 68.6 & 83.8 & 53.5 \\

        \bottomrule
    \end{tabular}
\end{table}

\subsection{Inference Speed and Response Length}\label{sec:app-speed}
Here, we report the improved inference efficiency of our model due to the concise nature of its responses in \autoref{tab:speed}. We use vLLM~\citep{kwon2023efficient} as our inference engine with a max model length of $16{,}384$ on a single H100-80GB GPU. Our model responds with $\approx$ $38$\% less time, $\approx$ $20$\% to $40$\% fewer tokens than the Monet-VLPO model on V* and MathVista, while achieving better performance, as detailed in \autoref{tab:monet}.
\begin{table}[t]
    \centering
    \caption{\textbf{Inference speed comparison. \textbf{Bold} entries denote that our model responds with fastest speed and least tokens.}}
    \label{tab:speed}
    \small
    \setlength{\tabcolsep}{6pt}
    \begin{tabular*}{\textwidth}{@{\extracolsep{\fill}} l cc cc @{}}
        \toprule
        \multirow{2}{*}{\textbf{Model}} 
        & \multicolumn{2}{c}{\textbf{V*}} 
        & \multicolumn{2}{c}{\textbf{MathVista}} \\
        \cmidrule(lr){2-3} \cmidrule(lr){4-5}
        & \text{avg. response time} & \text{avg. response length} 
        & \text{avg. response time} & \text{avg. response length} \\
        \midrule
        Monet-VLPO & $1.91$s & $52.85$ tokens & $4.43$s & $184.02$ tokens \\
        Ours     & $\mathbf{1.64}$s & $\mathbf{41.31}$ tokens & $\mathbf{2.74}$s & $\mathbf{111.02}$ tokens \\
        \bottomrule
    \end{tabular*}
\end{table}

\subsection{Controlling effects of response length}\label{sec:app-length}
Observing that our model responds with fewer tokens ($150$ tokens on average) than the VLPO model, we investigate whether penalizing longer response length alone can improve model performance. We experiment with penalizing $0.1$ to the responses with more than $150$ tokens. As shown in \autoref{tab:ablation_length_table}, this yields inferior performance, but can be partly mitigated by adding our attention reward, suggesting that controlling response length itself is not enough compared to our method design.
\begin{table}[H]
    \centering
        \captionof{table}{Using restricted response length with attention reward degrades performance compared with using attention reward alone. All models are trained for $400$ steps.}
        \label{tab:ablation_length_table}
    \vspace{8pt}
    \small
    \setlength{\tabcolsep}{3.5pt}
    \resizebox{0.8\textwidth}{!}{
        \begin{tabular}{@{}lLLLLL@{}}
            \toprule
            \multirow{1}{*}{\textbf{Reward Type}} 
            & \textbf{MathVista} 
            & \textbf{MMVP} 
            & \textbf{MMBench} 
            & \textbf{V*}
            & \textbf{HRBench4K} \\
            \midrule
            $-0.1$ if len > $150$
            & 62.5 & 73.3 & 49.1 & 80.6 & 68.0 \\
            $-0.1$ if len > 150 + $R_A$
            & 63.6 & 76.7 & 78.7 & 81.2 & 72.8 \\
            \bottomrule

    \end{tabular}
    }
\end{table}

\section{Inference Examples}\label{sec:app-examples}
We show more inference examples generated by the Monet-VLPO model and our model below.

\begin{figure*}[ht]
    \begin{tcolorbox}[
        colback=gray!3,
        colframe=black!35,
        title=\textbf{\small Inference Example 1: Chart Math Reasoning (MathVerse)},
        fonttitle=\bfseries,
        boxrule=0.6pt,
        arc=2pt,
        left=6pt,
        right=6pt,
        top=6pt,
        bottom=6pt
    ]

    \begin{minipage}[t]{0.32\textwidth}
    \vspace{0pt}
        \centering
        \includegraphics[width=\linewidth]{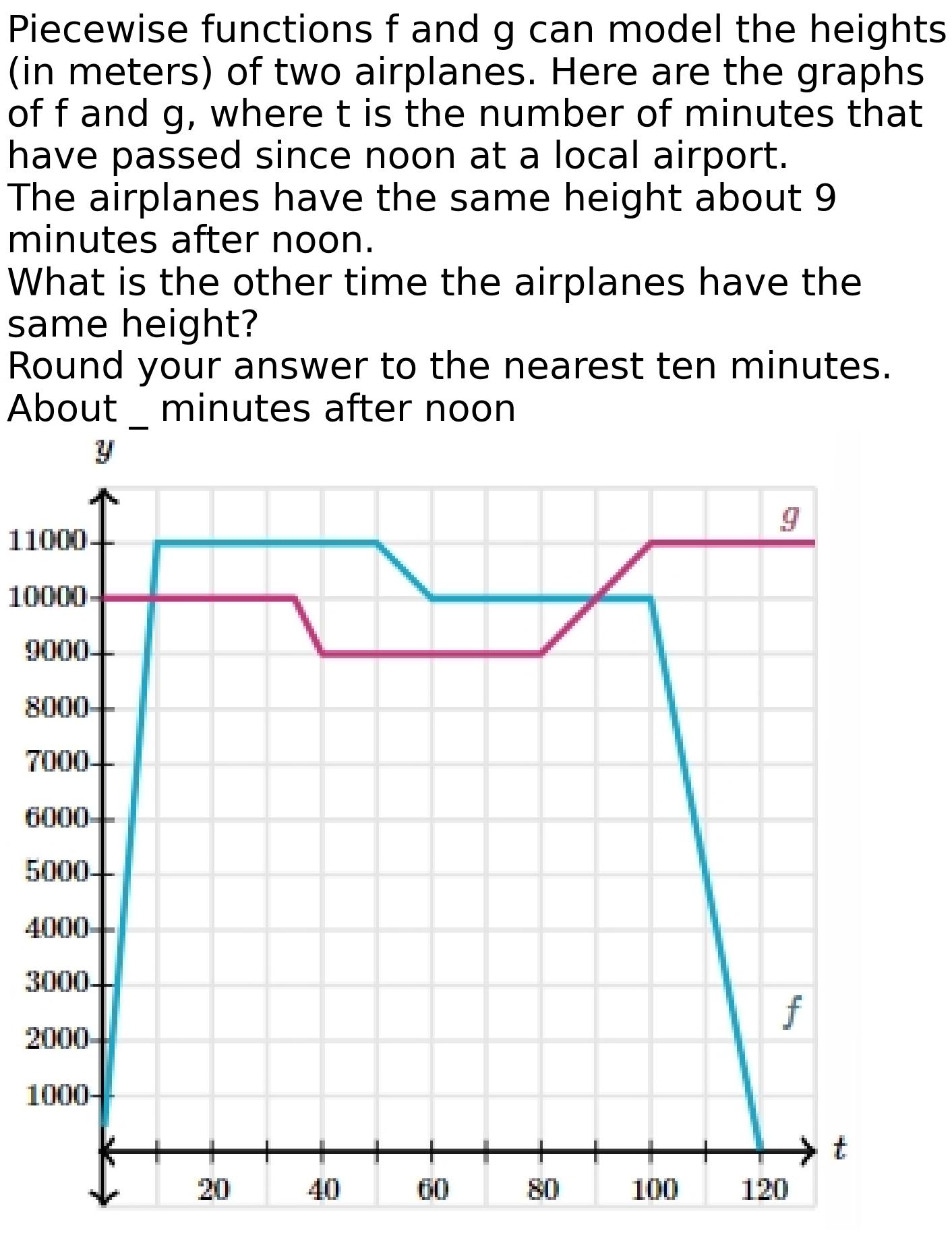}
        \small
        \textit{Input image}
    \end{minipage}
    \hfill
    \begin{minipage}[t]{0.64\textwidth}
    \vspace{0pt}
        \small
        \textbf{Question Prompt.}
Please directly answer the question and provide the final value, e.g., 1, 2.5, 300.

        \vspace{2pt}
        \textbf{Question:} Piecewise functions f and g can model the heights (in meters) of two airplanes. Here are the graphs of f and g, where t is the number of minutes that have passed since noon at a local airport.
The airplanes have the same height about 9 minutes after noon.
What is the other time the airplanes have the same height?
Round your answer to the nearest ten minutes.
About \_ minutes after noon \textbf{(Answer: 90)}

        \vspace{4pt}
        \textbf{Response by Monet-VLPO.}
        Find the other time the functions have the same height. The functions have the same height at the points where the functions intersect. The functions intersect at the points (9, 11000) and (119, 11000). 119 rounded to the nearest ten is 120. The other time the functions have the same height is 120 minutes after noon.

        \vspace{4pt}
        \textbf{Response by Ours.}
        The problem asks for the other time the airplanes have the same height, given that they have the same height about 9 minutes after noon. The provided image clearly shows that the two functions intersect again around the 99 mark on the horizontal axis. The problem specifies to round the answer to the nearest ten minutes. The final answer is $\boxed{\text{90}}$.

    \end{minipage}

    \end{tcolorbox}
    \caption{Monet-VLPO fails to locate the intersection of the two lines, while our model precisely locates the coordinates of the intersection.}
\end{figure*}

\begin{figure*}[b]
    \centering
    \vspace{-25pt}
    \begin{tcolorbox}[
        colback=gray!3,
        colframe=black!35,
        title=\textbf{\small Inference Example 2: Visual Perception and Spatial Reasoning (MME-RealWorld-Lite)},
        fonttitle=\bfseries,
        boxrule=0.6pt,
        arc=2pt,
        left=6pt,
        right=6pt,
        top=6pt,
        bottom=6pt
    ]

    \begin{minipage}[t]{0.32\textwidth}
    \vspace{0pt}
        \centering
        \includegraphics[width=\linewidth]{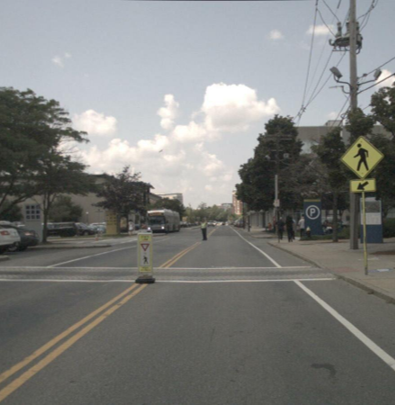}

        \small
        \textit{Input image}
    \end{minipage}
    \hfill
    \begin{minipage}[t]{0.64\textwidth}
    \vspace{0pt}
        \small
        \textbf{Question Prompt.}
        This image shows the front view of the ego car. What is the status of the buses that are to the front of the ego car?
        The choices are listed below. \textbf{(Answer: D)}

        \vspace{2pt}
        \textbf{(A)} One of the buses is moving, and two are stopped. \\
        \textbf{(B)} Three buses are stopped. \\
        \textbf{(C)} Many buses are stopped. \\
        \textbf{(D)} One of the buses is moving, and one is stopped. \\
        \textbf{(E)} The image does not feature the object.

        \vspace{4pt}
        \textbf{Response by Monet-VLPO.}
        First, I need to identify the buses in the image and determine their current status, moving or stopped. I will focus on the area in front of the ego car to get a clear view of the buses.
        \texttt{<latent\_start><latent><latent\_end>}
        The zoomed-in image clearly shows three buses in the distance. All three buses appear to be stationary and not in motion. The double yellow lines on the road also suggest that the area is a no-passing zone, which is consistent with the buses being stopped at a traffic light or a similar stationary point.
        \textbf{ANSWER:} The best answer is $\boxed{\text{B}}$.

        \vspace{4pt}
        \textbf{Response by Ours.}
        The image clearly shows one of the buses is moving, and one is stopped. The final answer is $\boxed{\text{D}}$.
    \end{minipage}

    \end{tcolorbox}
    \caption{Monet-VLPO responds with seemingly fluent reasoning but ends with hallucinations, while ours correctly identifies the one stopped bus.}
\end{figure*}

\begin{figure*}[t]

    \begin{tcolorbox}[
        colback=gray!3,
        colframe=black!35,
        title=\textbf{\small Inference Example 3: Geometry Math Reasoning (MathVerse)},
        fonttitle=\bfseries,
        boxrule=0.6pt,
        arc=2pt,
        left=6pt,
        right=6pt,
        top=6pt,
        bottom=6pt
    ]

    \begin{minipage}[t]{0.32\textwidth}
    \vspace{0pt}
        \centering
        \includegraphics[width=\linewidth]{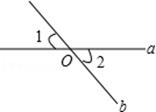}
        \small
        \textit{Input image}
    \end{minipage}
    \hfill
    \begin{minipage}[t]{0.64\textwidth}
    \vspace{0pt}
        \small
        \textbf{Question Prompt.}
        Please directly answer the question and provide the correct option letter, e.g., A, B, C, D.

        \vspace{2pt}
        \textbf{Question:} As shown in the figure, if angle 1 is equal to $50.0^\circ$, then angle 2 is equal to \underline{\hspace{1em}}. \textbf{(Answer: A)}

        \vspace{2pt}
        \textbf{A:} $50^\circ$ \quad
        \textbf{B:} $40^\circ$ \quad
        \textbf{C:} $140^\circ$ \quad
        \textbf{D:} $130^\circ$

        \vspace{4pt}
        \textbf{Response by Monet-VLPO.}
        To find the measure of angle 2, we can use the property that angles on a straight line sum up to $180^\circ$. Given that angle 1 is $50^\circ$, we calculate angle 2 as $180^\circ - 50^\circ = 130^\circ$. Hence, the measure of angle 2 is $130^\circ$. The correct answer is $\boxed{\text{D}}$.

        \vspace{4pt}
        \textbf{Response by Ours.}
        The problem asks for the measure of angle 2 given that angle 1 is $50^\circ$. The angles 1 and 2 are vertical angles, and vertical angles are always congruent. This means that angle 2 is also $50^\circ$. The final answer is $\boxed{\text{A}}$.
    \end{minipage}

    \end{tcolorbox}

    \caption{Monet-VLPO fails to capture the position of angle 2 and continue the calculation based on the wrong logic, while our model correctly identifies that angles 1 and 2 are congruent.}
\end{figure*}

\begin{figure*}[t]
    \centering
    \vspace{6pt}

    \vspace{-25pt}

    \begin{tcolorbox}[
        colback=gray!3,
        colframe=black!35,
        title=\textbf{\small Inference Example 4: Physical Attribute Reasoning (MMBench)},
        fonttitle=\bfseries,
        boxrule=0.6pt,
        arc=2pt,
        left=6pt,
        right=6pt,
        top=6pt,
        bottom=6pt
    ]

    \begin{minipage}[t]{0.32\textwidth}
    \vspace{0pt}
        \centering
        \includegraphics[width=\linewidth]{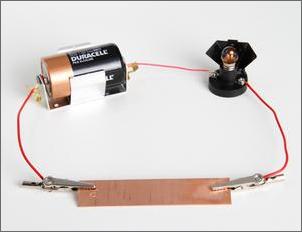}
        \small
        \textit{Input image}
    \end{minipage}
    \hfill
    \begin{minipage}[t]{0.64\textwidth}
    \vspace{0pt}
        \small
        \textbf{Question Prompt.}
        Identify the question that Gina's experiment can best answer.
        \vspace{2pt}
        
        \textbf{Question:} 
        The passage below describes an experiment. Read the passage and then follow the instructions below.
        
        Gina built an electric circuit: she used wires to connect a battery to a light bulb, the light bulb to a small piece of copper, and the copper back to the battery. When the circuit was complete, the light turned on. Gina observed the brightness of the light for five seconds. She then replaced the copper with a piece of iron of equal size and noted whether the light became brighter or dimmer. Gina built three more of the same type of circuit. She repeated the tests with each circuit. Gina recorded whether the circuits produced brighter light when the circuit included copper or when the circuit included iron.
        
        Figure: a circuit with a battery, a light bulb, and a piece of copper. 

        \textbf{(Answer: B)}
        
        \textbf{A}: Can light bulbs stay lit longer when circuits include copper or when circuits include iron? 
        
        \textbf{B}: Do circuits that include iron produce dimmer light than circuits that include copper?

        \vspace{4pt}
        \textbf{Response by Monet-VLPO.}
        The question asks what Gina's experiment can best answer. The passage and the hint indicate that Gina's experiment involves comparing the brightness of a light bulb in a circuit with copper versus a circuit with iron. The hint specifically mentions that Gina observed the brightness of the light for five seconds and noted whether the light became brighter or dimmer when the copper was replaced with iron. Therefore, the experiment can best answer the question of whether circuits that include iron produce dimmer light than circuits that include copper. The final answer is $\boxed{\text{B}}$.

        \vspace{4pt}
        \textbf{Response by Ours.}
        The problem asks for the question that Gina's experiment can best answer based on the provided passage and the figure. The passage clearly states that Gina observed the brightness of the light for five seconds and then replaced the copper with iron to note the change in brightness. The passage also mentions that she recorded whether the circuits produced brighter light when the circuit included copper or when the circuit included iron. This directly addresses the second option.
        ANSWER: The final answer is $\boxed{\text{B}}$.

    \end{minipage}

    \end{tcolorbox}

    \caption{Both Monet-VLPO and our model can reason correctly when the image does not provide relevant information for the question.}
    \vspace{15pt}
\end{figure*}

\begin{figure*}[t]
    \centering
    \vspace{6pt}

    \vspace{-25pt}

    \begin{tcolorbox}[
        colback=gray!3,
        colframe=black!35,
        title=\textbf{\small Inference Example 5: Coarse Perception (MMVP)},
        fonttitle=\bfseries,
        boxrule=0.6pt,
        arc=2pt,
        left=6pt,
        right=6pt,
        top=6pt,
        bottom=6pt
    ]

    \begin{minipage}[t]{0.32\textwidth}
    \vspace{0pt}
        \centering
        \includegraphics[width=\linewidth]{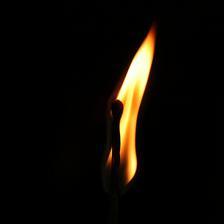}
        \small
        \textit{Input image}
    \end{minipage}
    \hfill
    \begin{minipage}[t]{0.64\textwidth}
    \vspace{0pt}
        \small
        \textbf{Question Prompt.}
        n/a
        \vspace{2pt}
        
        \textbf{Question:} 
        Is the flame of the match more round or thin? 
        
        \textbf{A.} Round
        
        \textbf{B.} Thin \textbf{(Answer: B)}
                
        \vspace{4pt}
        \textbf{Response by Monet-VLPO.}
        The flame of the match is  more  round, as it appears wider and less elongated.  Therefore, the final answer is $\boxed{\text{A}}$.

        \vspace{4pt}
        \textbf{Response by Ours.}
        ANSWER: The flame of the match is thin, so the correct answer is $\boxed{\text{B}}$.

    \end{minipage}

    \end{tcolorbox}

    \caption{Monet-VLPO fails to capture the exact shape of the flame in the image using pure text reasoning, while our model correctly recognizes the flame’s thin shape.}
\end{figure*}

\begin{figure*}[t]
    \centering
    \vspace{6pt}

    \vspace{-25pt}

    \begin{tcolorbox}[
        colback=gray!3,
        colframe=black!35,
        title=\textbf{\small Inference Example 6: Algebraic Reasoning (MathVista)},
        fonttitle=\bfseries,
        boxrule=0.6pt,
        arc=2pt,
        left=6pt,
        right=6pt,
        top=6pt,
        bottom=6pt
    ]

    \begin{minipage}[t]{0.32\textwidth}
    \vspace{0pt}
        \centering
        \includegraphics[width=\linewidth]{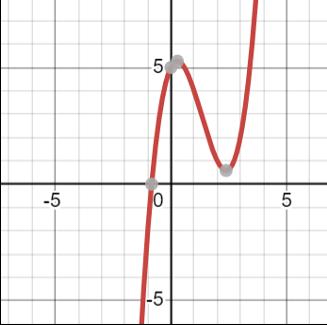}
        \small
        \textit{Input image}
    \end{minipage}
    \hfill
    \begin{minipage}[t]{0.64\textwidth}
    \vspace{0pt}
        \small
        \textbf{Question Prompt.}
        Hint: Please answer the question and provide the correct option letter, e.g., A, B, C, D, at the end.

        \vspace{2pt}
        
        \textbf{Question:} 
        
        Question: is f(3) > 0?
        
        Choices:   
        
        \textbf{A.} Yes
        
        \textbf{B.} No 
        
        \textbf{(Answer: A)}
                
        \vspace{4pt}
        \textbf{Response by Monet-VLPO.}
        To determine if f(3) is greater than 0, we need to locate the point where x = 3 on the given graph and observe its position relative to the x-axis.
        The point where x = 3 is located on the graph. Observing the position of this point, we can see that it is above the x-axis, indicating a positive value for f(3).
        THOUGHT N: The point where x = 3 is above the x-axis, indicating a positive value for f(3).
        ANSWER: The final answer is $\boxed{\text{A}}$.

        \vspace{4pt}
        \textbf{Response by Ours.}
        ANSWER: The problem asks for the value of the function f at x=3 and whether it is greater than 0. The provided image is a graph of the function f. I need to locate the point on the graph where x=3 and then observe the corresponding y value to answer the question.Therefore, the final answer is $\boxed{\text{No}}$.

    \end{minipage}

    \end{tcolorbox}

    \caption{Our model may fail to produce the correct final answer despite demonstrating correct reasoning toward a solution, which could be a hallucination inherited from the base model.}
\end{figure*}

\begin{figure*}[t]
    \centering
    \vspace{6pt}

    \vspace{-25pt}

    \begin{tcolorbox}[
        colback=gray!3,
        colframe=black!35,
        title=\textbf{\small Inference Example 7: Spatial Reasoning (HRBench4K)},
        fonttitle=\bfseries,
        boxrule=0.6pt,
        arc=2pt,
        left=6pt,
        right=6pt,
        top=6pt,
        bottom=6pt
    ]

    \begin{minipage}[t]{0.32\textwidth}
    \vspace{0pt}
        \centering
        \includegraphics[width=\linewidth]{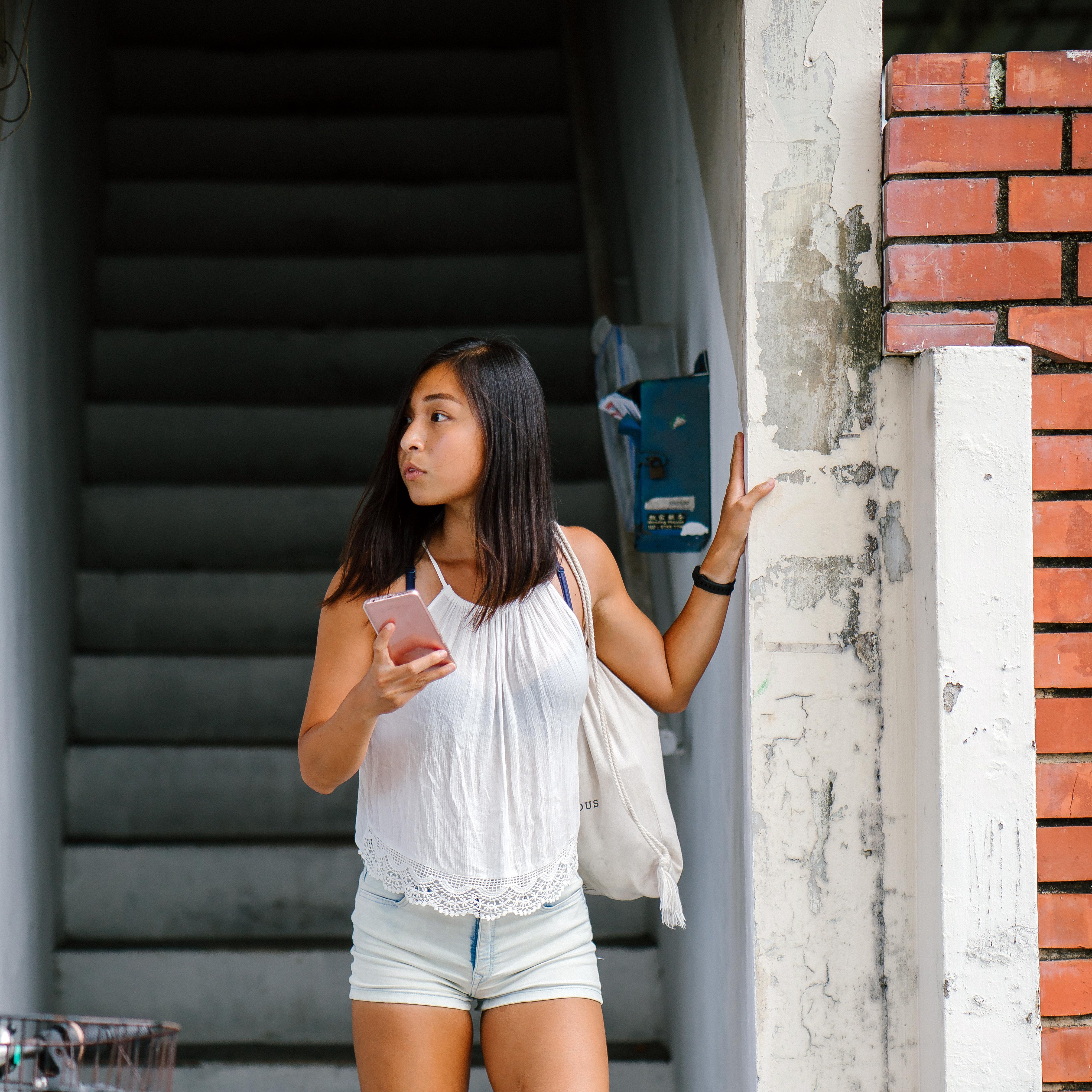}
        \small
        \textit{Input image}
    \end{minipage}
    \hfill
    \begin{minipage}[t]{0.64\textwidth}
    \vspace{0pt}
        \small
        \textbf{Question Prompt.}
        n/a
        \vspace{2pt}
        
        \textbf{Question:} 
        
        Is the mailbox on the left or right side of the woman?
        
        \textbf{A.} Directly above
        
        \textbf{B.} In front 
           
        \textbf{C.} Left
                
        \textbf{D.} Right
        
        \textbf{(Answer: D)}
                
        \vspace{4pt}
        \textbf{Response by Monet-VLPO.}
        To determine the relative position of the mailbox to the woman, I will analyze the image.
        \texttt{<latent\_start><latent><latent\_end>}

        The image clearly shows the  mailbox  is located  to the  right of the woman,  next to the pillar she is leaning on. The woman is  on the left side of the mailbox.
        THOUGHT N: The mailbox is on the right side of the woman.Therefore, the final answer is $\boxed{\text{D}}$.

        \vspace{4pt}
        \textbf{Response by Ours.}
        ANSWER: The image clearly shows the mailbox is positioned directly above the woman's head, relative to the viewer's perspective. The options provided are A. Directly above, B. In front, C. Left, and D. Right. The correct answer is $\boxed{\text{A}}$.

    \end{minipage}

    \end{tcolorbox}

    \caption{Our model may fail to perceive the correct spatial relationship between objects, likely due to conflating the target object's detection with the taller box behind the mailbox.}
\end{figure*}


\end{document}